%% file: main.tex
\documentclass[sigconf,preprint]{acmart}
\usepackage{algorithm}  
\usepackage{algpseudocode}  
\usepackage{amsmath}

\AtBeginDocument{%
  }

\usepackage{wrapfig}
\usepackage{subcaption}
\usepackage{balance} 

\usepackage{microtype}
\usepackage{graphicx}
\usepackage{booktabs} % for professional tables
\usepackage{xcolor}
\usepackage{bm,bbm}
\usepackage{caption}
\usepackage[utf8]{inputenc} % allow utf-8 input
\usepackage[T1]{fontenc}    % use 8-bit T1 fonts
\usepackage{hyperref}       % hyperlinks
\usepackage{url}            % simple URL typesetting
\usepackage{booktabs}       % professional-quality tables
\usepackage{amsfonts}       % blackboard math symbols
\usepackage{nicefrac}       % compact symbols for 1/2, etc.
\usepackage{microtype}      % microtypography
\usepackage{color}  
\usepackage{amsmath, amsthm, multirow, paralist}
\usepackage{mathtools}
\usepackage{bm,bbm}
\usepackage{hyperref}
\hypersetup{colorlinks=true,breaklinks=true}

\usepackage{algorithm}
\usepackage{listings}
\newtheorem*{theorem*}{Theorem}
\title{Sparse-VQ Transformer: An FFN-Free Framework with Vector Quantization for Enhanced Time Series Forecasting}

\author{Yanjun Zhao}
\authornote{Two authors contributed equally to this research.}
\email{yanjun.zhao@stu.xjtu.edu.cn}
\affiliation{%
  \institution{Xi'an Jiaotong University, Alibaba Group}
  \country{China}
}

\author{Tian Zhou}
\authornotemark[1]
\email{tian.zt@alibaba-inc.com}
\affiliation{%
  \institution{Alibaba Group}
  \country{China}
  \postcode{}
}

\author{Chao Chen}
\email{cc410784@alibaba-inc.com}
\affiliation{%
  \institution{Alibaba Group}
  \country{China}
}

\author{Liang Sun}
\email{liang.sun@alibaba-inc.com}
\affiliation{%
  \institution{Alibaba Group}
  \country{USA}
}

\author{Yi Qian}
\email{yqian@mail.xjtu.edu.cn}
\affiliation{%
  \institution{Xi'an Jiaotong University}
  \country{China}
}

\author{Rong Jin}
\authornote{The author now works at Meta Platforms, Inc.}
\email{rongjinemail@gmail.com}
\affiliation{%
  \institution{Alibaba Group}
  \country{USA}
}

\renewcommand{\shortauthors}{Yanjun Zhao et al.}
\begin{document}

  \begin{abstract}

Time series analysis is vital for numerous applications, and transformers have become increasingly prominent in this domain. Leading methods customize the transformer architecture from NLP and CV, utilizing a patching technique to convert continuous signals into segments. Yet, time series data are uniquely challenging due to significant distribution shifts and intrinsic noise levels. To address these two challenges,
we introduce the Sparse Vector Quantized FFN-Free Transformer
(Sparse-VQ). Our methodology capitalizes on a sparse vector quantization technique coupled with Reverse Instance Normalization (RevIN) to reduce noise impact and capture sufficient statistics for forecasting, serving as an alternative to the Feed-Forward layer (FFN) in the transformer architecture. Our FFN-free approach trims the parameter count, enhancing computational efficiency and reducing overfitting. Through evaluations across ten benchmark datasets, including the newly introduced CAISO dataset, Sparse-VQ surpasses leading models with a \textbf{7.84\%} and \textbf{4.17\%} decrease in MAE for univariate and multivariate time series forecasting, respectively. Moreover, it can be seamlessly integrated with existing transformer-based models to elevate their performance. Our source code and the new dataset are available at: https://anonymous.4open.science/r/Sparse-VQ-DC28.

  \end{abstract}

%    <concept>
%        <concept_id>10010147.10010257.10010293.10010294</concept_id>
%        <concept_desc>Computing methodologies~Neural networks</concept_desc>
%        <concept_significance>500</concept_significance>
%        </concept>
%  </ccs2012>

% \ccsdesc[500]{Computing methodologies~Neural networks}



\maketitle

%\footnote{$*$ contributed equally to this work}
\input{sections/1_introduction}

%\input{sections/2_related_work}
\input{sections/2_related_work_concise}
\input{sections/7_sparse_representation}

\input{sections/3_methods}

\input{sections/4_experiments}

\input{sections/5_conclusions}
\bibliographystyle{ACM-Reference-Format.bst}
\balance
\newpage
\bibliography{6_mybib}

%%%%%%%%%%%%%%%%%%%%%%%%%%%%%%%%%%%%%%%%%%%%%%%%%%%%%%%%%%%%
\newpage
\appendix
\input{sections/6_Appendix}

\end{document}

%% file: sections/1_introduction.tex
\section{Introduction}\label{sec_intro}

%{\color{blue}

Time series forecasting involves making predictions based on historical data, which is widely used in various real-world applications including weather forecasting, stock prediction, energy consumption planning, E-commerce supply chain scheduling, etc. With the evolution of deep learning techniques, traditional statistical approaches~\cite{box_arima2,Holt-Winter18,DBLP:journals/corr/FlunkertSG17-deepAR,Taylor2018ForecastingAS} have been largely superseded by deep learning models~\cite{TCN,dual-state-attention-rnn-qin}. More recently, the success of Transformers in the NLP and CV domain~\cite{attention_is_all_you_need,devlin-etal-2019-bert,dosovitskiy2021an,rao2021global} has led to their adoption in time series forecasting tasks and yielded promising results, making a significant shift in the techniques employed for predictive analytics.

\begin{figure}[h]
%\vskip -0.15in
\centering
% \scalebox{0.70}{
\includegraphics[width=0.99\linewidth]{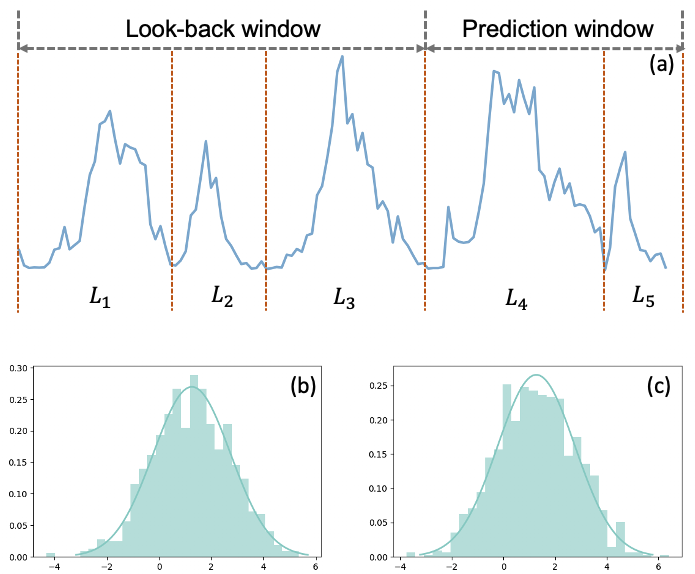}
\caption{Although the temporal covariate shift problem exists in non-stationary time series, as shown in (a), historical patterns may still reoccur in the future. For example, the distribution of $L_{2}$ (b) are similar to that of $L_{5}$ (c).}
\label{fig:distribution_shift}
%\vskip -0.15in
\end{figure}

However, a significant portion of time series data generated in real-world scenarios inherently exhibits non-stationary property. This suggests that statistical features, including low-order statistics like the mean and variance, as well as high-order statistics encompassing higher moments, often experience variations over time, a phenomenon referred to as distribution shift~\cite{fan2023dish}. This issue often leads to poor generalization, posing significant challenges in time series forecasting. Previous studies~\cite{Adaptive_Input_Normalization,Adaptive_Normalization,fan2023dish} focused on applying normalization as a pre-processing technique for time series prediction, which mitigates the non-stationarity of the raw time series and offers a relatively stable data distribution for models, resulting in improved predictability. Building on these, RevIN~\cite{kim2022reversible} proposed to restore the low-order statistical information of a time-series instance and implement the reversible normalization after the model output, which has been proven powerful and used extensively. Another common challenge encountered in time series analysis is the presence of high noise or low signal density. When the vanilla transformer model is applied to individual time points, its performance tends to be unsatisfactory. Recent studies showed that this issue can be mitigated by employing a straightforward patching technique to tokenize the signal, thereby enhancing its representation~\cite{patchTST,lin2023petformer}.

% Based on the aforementioned difficulties, we propose our Sparse-VQ framework. which maps long-range time series to the discrete tokens and replaces the original data by the learnable codebook based on distance, forming a discrete space. In comparison to the continuous latent embedding space, the model can effectively denoise the data and decrease the variance. What's more, distinct tokens represent different patterns that have occurred in the past. It is worth affirming that historical patterns still hold undeniable value because they have the potential to reoccur in similar forms in the future, as shown in Figure \ref{fig:distribution_shift}. Previous studies (Geva et al., 2021\cite{geva-etal-2021-transformer}, Dai et al., 2022 \cite{Knowledge_Neurons}) in the field of Natural Language Processing have demonstrated that the FFN (Feed-Forward Network) within the attention mechanism primarily functions as memory. Building on these findings, we conducted further investigations into the role of this structure in the field of time series forecasting. Our findings revealed that the FFN structure aids in preserving crucial distribution information of the data, including measures such as mean and variance. As the utilized RevIN method pre-stores mean and variance information in the model, the FFN structure becomes superfluous. As a result, we introduce a Feed-Forward Network (FFN)-free architecture that incorporates Reversible Instances (RevIn) to directly model these statistics. 
  
Transformers have made significant strides in NLP and image recognition, with applications such as ChatGPT and Midjourney showcasing their capabilities. Studies reveal that a single attention layer in transformers can be seen as a compound of bigram and "skip-trigram" (patterns like "A… B C") configurations. Each attention head can deftly navigate from a given token ("B") to an antecedent one ("A"), influencing the likelihood of a successive token ("C")—a key engineering feat driving transformer success. Furthermore, the Feed-Forward Network (FFN) module serves as a conceptual archive, encoding token co-occurrences (A, B, C)—indicative of high-order statistics~\cite{Geva2020TransformerFL, geva-etal-2021-transformer, Knowledge_Neurons}.

In contrast, time series data are sequential numerical recordings, distinct from NLP tokens or vision patches. Reliable extraction of statistical features from noisy time series is crucial for accurate forecasting. However, FFNs encounter limitations in modeling simple polynomial relationships (e.g., $x_i * x_j$) due to MLPs' inherent approximation challenges~\cite{TAO2020102076}. Consequently, it remains an open question whether the conventional FFN-plus-attention architecture of transformers is optimal for time series forecasting.

%Thus, the deployment of the FFN module in transformer implies it is preferred by scenarios with more or less stationary distributions and stable statistics. 

Given that time series data often exhibit drift in data distribution, we expect a significant change over time in low-order statistics mean, variance and high order statistics, a less desirable case for the usage of the FFN module.  
In our study, we develop a hybrid approach, dubbed {\bf Sparse-VQ}, that can better capture statistics of drifting distributions than the FFN module: we first apply Reversible Instance Normalization (RevIN) to directly model local low-order statistics, specifically the mean and variance, and then employ vector quantization to reduce noise from inputs and capture global statistics. Our extensive empirical studies show that Sparse-VQ can substantially improve the prediction performance for non-stationary and noisy signals over the FFN module, leading to what we call FFN-free transformer for time series forecasting. This FFN-free approach significantly reduces the model's parameter count, leading to an significant improvement in computation and improved generalization. 
% an alleviation of the overfitting problem from transformer. 

Here we summarize our key contributions as follows:
\begin{enumerate}
    \item We propose the Sparse-VQ structure that embeds long time series into a discrete space, thus effectively reducing the impact of noise from the data. 
    \item We investigate the effect of FFN structure in Transformer and propose a FFN-free Transformer structure that can achieve a remarkable 21.52\% reduction in parameters with improved performance.
    \item We have conducted extensive experiments on ten diverse benchmark datasets, including the novel CAISO dataset, and the periodic Traffic dataset, which is ideal for testing forecasting periodic time series. 
    % Its distinct periodicity, reminiscent of the Traffic dataset, serves as an ideal testbed to showcase our model's adeptness at forecasting periodic signals. 
    Our empirical studies show that compared with state-of-the-art methods, Sparse-VQ can reduce the prediction error  by 4.17\% and 7.84\% for multivariate and univariate forecasting, respectively. Also, our empirical findings show that the proposed framework has the potential to substantially enhance the performance of all Tranformer-based models.
    % \item We introduce the CAISO dataset, tailored for time series forecasting. Visual analysis reveals that CAISO exhibits a pronounced periodic pattern akin to the Traffic dataset, distinguishing it from other datasets. This characteristic makes it an ideal benchmark to highlight the model's proficiency in predicting periodic signals.
\end{enumerate}

The remainder of this paper is structured as follows: In Section~\ref{sec_related_works_concise}, we provide a summary of related work. Section~\ref{sec_methods} presents the detailed framework designed for our proposed approach and introduces the Sparse-VQ and FFN-free Transformer structure. In Section~\ref{sec_experiments}, we conduct the numerical experiments to evaluate the performance of our proposed method in long-time, short-time and few-shot forecasting tasks compared to various SOTA baseline models. Furthermore, we present ablation experiments, boosting results of Sparse-VQ, different variants of Vector Quantization and robustness studies. Finally, in Section ~\ref{sec_conclusion}, the conclusions and future research directions are discussed.

%end color blue
%}

%% file: sections/2_related_work_concise.tex
\begin{figure*}[t]
\centering
%\setlength{\abovecaptionskip}{0.2cm}
% \scalebox{0.70}{
\includegraphics[width=0.9\linewidth]{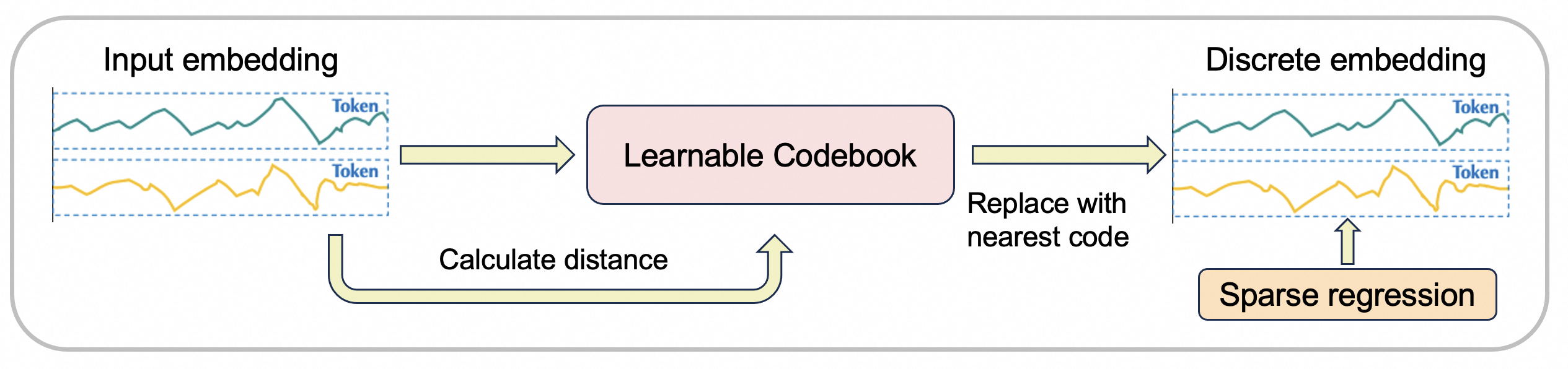}
\caption{Sparse-VQ Block(SVQ).}
\label{fig:SVQ}
% \vskip -0.2in
\end{figure*}

\begin{figure*}[t]
\centering
%\setlength{\abovecaptionskip}{0.2cm}
%\scalebox{0.80}{
\includegraphics[width=0.85\linewidth]{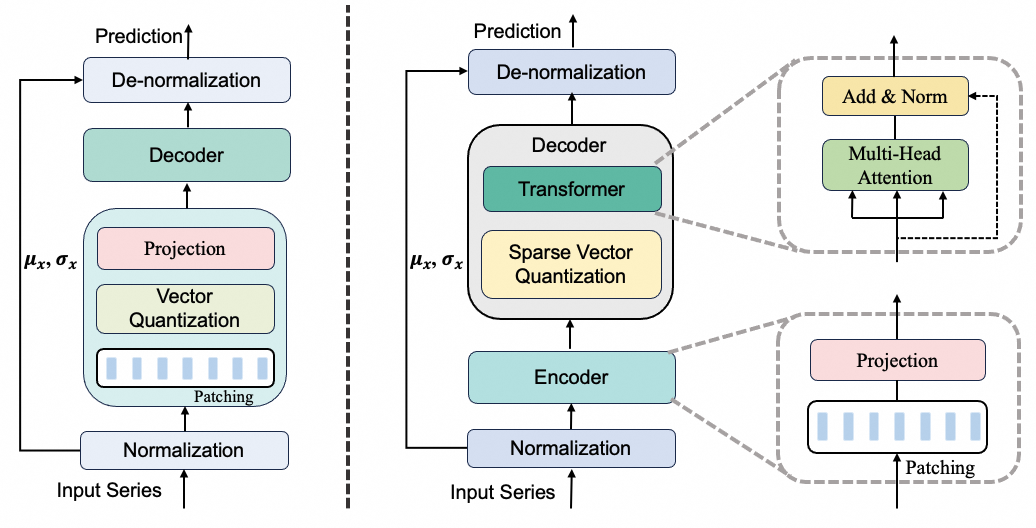}
\caption{Classic model with incorporation of VQ(left) and Sparse-VQ Model Overview(right).}
\label{fig:model_structure}
%\vskip -0.2in
\end{figure*}

\section{Related Work}
\label{sec_related_works_concise}
Here we present a concise summary of related work on time series forecasting, distribution shift, and vector quantization. For a more detailed review, please refer to Appendix \ref{sec_appendix_related_works}.

\subsection{Time Series Forecasting}
Time series forecasting has progressed from traditional statistical algorithms like ARIMA~\cite{box_arima2} and Holt-Winter~\cite{Holt-Winter18} to machine learning models including DeepAR~\cite{DBLP:journals/corr/FlunkertSG17-deepAR} and Prophet~\cite{Taylor2018ForecastingAS}. The advent of deep learning has led to the adoption of sophisticated neural networks such as TCN~\cite{TCN} and LSTM~\cite{hochreiter_long_1997_lstm}, with notable contributions from N-BEATS~\cite{nbeats}, which excelled in the M4 competition. Transformer-based models, including Informer~\cite{haoyietal-informer-2021}, Reformer~\cite{KitaevKL20-reformer}, and Autoformer~\cite{Autoformer}, have significantly improved time series forecasting. Innovations like TimesNet~\cite{TIMESNET}, FiLM~\cite{zhou2022film}, FEDformer~\cite{FedFormer}, and Pyraformer~\cite{liu2022pyraformer} continue to refine these techniques. Pretrained models such as FPT~\cite{zhou2023onefitsall} showcase the adaptability of language or vision models to time series tasks, cementing Transformer-based methods as a top-tier approach~\cite{Wen2022transformers, Shao2023exploring}.

\subsection{Distribution Shift}
Addressing distribution shifts in non-stationary time series data is a persistent challenge. To tackle this issue, researchers mainly utilize domain adaptation~\cite{Tzeng17Domain_Adaptation, Ganin16domain_adaptation, Wang18Domain_Adaptation} and generalization strategies~\cite{Wang2021GeneralizingTU, Li18DomainGeneralization, pmlr-v28-muandet13, Lu2022DomaininvariantFE} to mitigate the distribution shift. Adaptive RNNs~\cite{Du2021AdaRNNAL}, RevIN~\cite{reversible}, and DIVERSIFY~\cite{lu2023outofdistribution} have emerged as novel solutions, leveraging period segmentation, normalization techniques, and out-of-distribution learning to cope with varying distributions.

\subsection{Vector Quantization}
Vector quantization (VQ) has been a key compression strategy in signal and image processing, with VQ-VAE~\cite{VQVAE} combining it with variational autoencoders for discrete and learnable priors. Developments such as SoundStream's residual VQ~\cite{residualVQ}, VQGAN~\cite{VQGAN}, and its extensions ViT-VQGAN~\cite{VQGAN_use_cosine_sim} and TE-VQGAN~\cite{orthogonal} have pushed VQ into new domains. Recent research has demonstrated the potential of applying Vector Quantization (VQ) to time series forecasting~\cite{Rasul2022VQARVQ}. However, our comprehensive study suggests that the conventional approach of pre-quantization may not enhance performance. Instead, implementing Vector Quantization after the enrichment of the signal by the encoder proves to be more beneficial.

%Our framework leverages VQ to enhance transformer capabilities in time series forecasting, opening new avenues for discrete tokenization of series data.

%% file: sections/7_sparse_representation.tex
\section{Sparse VQ representation}
\label{sec:eff_sparse_reg}

To differentiate the performance of the sparse regression technique from that of the traditional cluster-based approach in quantization, we evaluate the minimal count of codewords, denoted by $N(\mathcal{U}, \epsilon)$, required to represent any vector within a unit sphere $\mathcal{U}$ with an approximation error smaller than $\epsilon$. The following proposition illustrates that with sparse regression, $N(\mathcal{U}, \epsilon)$ can be significantly reduced from $O(1/\epsilon^n)$ to $O(1/\epsilon^q)$, where $q \ll n$ for vectors in a high-dimensional space.

\begin{proposition}
\label{proposition}
For a cluster-based scheme, $N(\mathcal{U}, \epsilon)$ is no less than $1/\epsilon^n$, whereas for the sparse regression technique, $N(\mathcal{U},\epsilon)$ has an upper bound of $(4n/\epsilon)^q$, where
\[
q \geq \max\left(3, \frac{\log(4/\epsilon)}{\log\log(2n/\eta)}\right),
\]
provided that the count of non-zero coefficients used in sparse regression is at least
\[
    \frac{4n}{\epsilon \left(\log K + q\log (4n) - (q+1)\log \epsilon \right)}.
\]
\end{proposition}
Due to the space limitation, we postpone the proof in Appendix~\ref{app:proof}. 

%%%Due to page constraints, the proof has been relegated to Appendix~\ref{app:proof}, which elaborates on the derivation based on the work ~\cite{chen2023svq}.

%% file: sections/3_methods.tex
\section{Method}\label{sec_methods}

In this section, we begin with the overall framework, as shown in Figure \ref{fig:model_structure}, and then follow by the detailed description of the Sparse-VQ and FFN-free Transformer structure.

\subsection{Overall Architecture}

Our proposed architecture consists of three components: encoder, Sparse-VQ, and decoder.  Initially, the input time series undergo normalization via the Reversible Instance Normalization (RevIN) technique~\cite{kim2022reversible}, followed by segmentation into patches. These patches are then projected onto a higher-dimensional space within the encoder to extract salient time series features. Subsequently, the SVQ module receives these features, mapping them onto a latent discretized space for discrete representation learning. The resulting discrete tokens are fed into the decoder, which operates on an FFN-free transformer framework. To finalize the forecasting process, we apply the inverse of the normalization procedure. We will further delineate the specifics of each component in the subsequent sections.

%\input{tables_vq/different_vq_structure}
\input{tables_vq/different_vq_strcture_concise}
\subsection{Sparse-VQ}

%{\color{red} Can we change the presentation of this subsection? }

In the pursuit of advancing time series forecasting, we propose a novel design framework, Sparse-VQ, described in Algorithm \ref{algo_SVQ} and visually represented in Figure \ref{fig:SVQ}. Sparse-VQ is innovative in its approach to reconstructing the original vector through a sparse combination of its nearest neighbors. This method is designed to finely tune the balance between noise reduction and signal preservation, a pivotal aspect of sparse regression that is particularly suited to the requirements of temporal data.

The inception of Sparse-VQ was prompted by the recognized shortcomings of conventional Vector Quantization (VQ) methods in the realm of time series forecasting. Our initial experiments with a standard model architecture, placing VQ before the encoder as shown in Figure \ref{fig:model_structure} (left), did not yield the expected enhancement in predictive accuracy, as evidenced by the results in Table \ref{tab:ablation_EncoderVQ}. The primary reason for this shortfall was the high noise-to-signal ratio typical of time series data, a stark deviation from what is commonly encountered in natural language processing. Upon further analysis, insights from studies like PatchTST~\cite{patchTST} and PETformer~\cite{lin2023petformer} confirmed our observations. These works highlight that augmenting the information density through patching techniques can significantly boost the efficacy of transformer-based models.

Motivated by these insights, we refined our methodology. Our approach allows the encoder to first distill a rich feature set from time series data before applying VQ. By doing so, VQ operates on more complex and variable embeddings, effectively mitigating variance with minimal impact on bias. This empowers our model to leverage the full potential of VQ towards improving overall prediction accuracy.

Further exploring the capabilities of VQ, we conducted a series of experiments to evaluate the impact of diverse VQ configurations: (a) $SVQ$: Sparse-VQ; (b) $VQ$: traditional Vector Quantization implemented in the decoder; (c) $VQ_{cosine}$: Vector Quantization using cosine similarity to measure distances; (d) $VQ_{K-means}$: Vector Quantization with a codebook initialized by K-means clustering centroids; (e) $VQ_{recursive}$: employing multiple vector quantizers recursively to quantize the residuals of layer outputs, as shown in Figure \ref{fig:VQ_recursive}; and (f) $VQ_{AdaptiveCodebook}$: Vector Quantization with an adaptive codebook learned via sparse regression, detailed in Figure \ref{fig:VQ_adaptive}. The empirical findings from these experimental studies are detailed in Table \ref{tab:different_vq_structure_concise}.

After analyzing the limitations of the standard VQ approach, we introduce Sparse-VQ as the focal point of our research, while also conducting comprehensive evaluations of various VQ design alternatives.
 
% Below, we introduce the Vector Quantization (VQ) technique into time series forecasting. Initially, our investigation embarked on an exploration of the conventional model structure with VQ, specifically by integrating VQ ahead of the model encoder, as shown in Figure \ref{fig:model_structure} (left). 

\input{tables_vq/EncoderVQ}

% The empirical outcomes in Table \ref{tab:ablation_EncoderVQ} show that the incorporation of VQ doesn't not lead to a stable improvement in prediction performance. This is primarily attributed to the relatively low information density characteristic of time series. It can be also corroborated from PatchTST~\cite{patchTST} and PETformer~\cite{lin2023petformer}, the latter has incorporated patching method to increase the information density of data, resulting in significant performance enhancements compared with other transformer-based models. %, thereby necessitating targeted improvements that cater to this characteristic. 
% As a result, while the VQ structure contributes to a variance reduction, it concurrently exerts an influence on the bias of the predictive results, causing an final decrement in the overall prediction accuracy.   

\begin{algorithm}[h]  
  \caption{Sparse-VQ}  
  \label{algo_SVQ}  
  \begin{algorithmic}[1]
    \Require  
        $y=Encoder(x)$, the output of the encoder;
    \Ensure 
 	The quantized  $\hat{y}$;
        \State Randomly initialize a leanrnable codebook $Z \in R^{D \times C}$;
        \State Calculate Euclidean Distance between the embedding $y$ and $Z$;
        \State Replace the embedding $y$ with its nearest neighbor in the codebook to reconstruct the discrete $\hat{y}$;
        \State $\hat{y} = SparseRegression( \hat{y} )$;
  \end{algorithmic}  
\end{algorithm}

\subsection{FFN-free Transformer}

\input{tables_vq/PatchTST_woReverseNorm}

We have opted to integrate a FFN-free architecture into our model. It is inspired by research of the effects of Feed-Forward networks (FFN) in natural language processing~\cite{ffn_transformer}, which suggests that FFN serves as key-value memories, enabling the preservation of contextual information in datasets with significant scale. In the context of language modeling, key-value memory is essentially to memorize co-occurrence of different tokens (e.g. skip grams~\cite{Geva2020TransformerFL}), a special form of high order statistics from training data, implying that the main role of the FFN module is to compute and store the data statistics for prediction. 
Following this speculation of the FNN module, we argue if it is appropriate to use the FFN module for non-stationary distribution where statistics of different orders vary significantly over time. 
%%%It is this speculation of the FNN module that makes us question if it is the most appropriate to use the FFN module for non-stationary distribution where statistics of different orders vary significantly over time. 

%Key-Value Memory Network~\cite{MN} also verifies this conclusion, which can be expressed as:

%\begin{equation}
%    MN(x) = softmax(X K ^ T)V,
%\end{equation}
%At the same time, Feed-forward networks can be expressed as :
%\begin{equation}
%    FFN(x) = Relu(X{W_1}^T)W_2,
%\end{equation}
%where $K,T,W_1,W_2 \in \R ^{d_m \times d} $.
\input{tables_vq/parameters}

Thus, we first investigate if the FFN module aids in retaining statistical measures (e.g. mean and variance) in the realm of time series forecasting. To this end, we conduct experiments of time series forecasting without using RevIN~\cite{kim2022reversible}, which is used to normalize time series data by local mean and variance. According to Table 3, we observe that, in the case of not using RevIN for data normalization, introducing FFN does significantly improve prediction accuracy, partially validating the hypothesis that the FFN module is used to capture the data statistics. Since both RevIN and FFN capture data statistics, it is thus redundant to include two different modules for the same role. In addition, since RevIN captures local mean and variance, it is more suitable for non-stationary distribution. This is in contrast to the FNN module, where static statistics are stored. In fact, our ablation study in Table ~\ref{tab:ablation_modules} shows that when using RevIN for data normalization, removing FFN from transformer can enhance prediction performance, partially validating the hypothesis that RevIN and FFN play similar, or sometimes even conflicting (i.e. stationary vs. drifting distributions) roles in time series forecasting.

%During the framework of PatchTST\cite{patchTST}, RevIN\cite{kim2022reversible} was employed, resulting in the pre-computation and storage of these statistical measures. Consequently, the inclusion of FFN is redundant and removing it can actually enhance the prediction performance. On the contrary, with the absence of ReverseNorm, FFN assumes a crucial role in information retention and in this scenario, retaining FFN improves prediction performance in comparison to its removal. Relative validation experiment results are shown in Table \ref{tab:PatchTST_woReverseNorm}.
%To substantiate above thesis, we carried out a series of validation experiment results are shown in Table \ref{tab:PatchTST_FFN_topic}

%By removing the FFN module, our network can effectively reduce the number of parameters by an average of 21.52\%, as detailed in Table \ref{tab:FFN_parameters}.

Moreover, within the time series domain, which typically features data of relatively low-rank dimensions (e.g., the seasonality of time series can be considered as a low-rank structure), the risk of overfitting is exacerbated. Our FFN-free architecture addresses this issue by effectively reducing the number of parameters by an average of 21.52\%, as detailed in Table \ref{tab:FFN_parameters}. %This reduction in model complexity serves as a safeguard against overfitting.
The reduction of model complexity further justifies the removal of FFN module by improving both computational efficiency and model generalization.

\subsection{Optimization}

\textbf{Prediction Loss.} We utilize the MAE loss to measure the discrepancy between the prediction and the ground truth, which can be written as:

\begin{equation}
    \mathbb{L}_{pred} = \frac{1}{M} \sum_{i=1}^{M} { \frac{1}{T} \sum_{j=1}^{T} \left| \hat{x}_{L+j} ^ {(i)} - x_{L+j} ^ {(i)} \right| },
\end{equation}
where $M$ is the number of channels of the time series and $T$ is the prediction length.

\textbf{Commitment Loss.} We also add the commitment loss to promote the proximity of the input to the selected codebook vector and minimize variations among codebook embeddings, thereby ensuring that the input does not frequently switch between different codebook embeddings. The commitment loss is formulated as follows:

\begin{equation}
    \mathbb{L}_{ct} = {\Vert sg[x] - vq(x) \Vert} _2 ^2 + {\Vert x - sg[vq(x)] \Vert} _2 ^2,
\end{equation}
where $sg$ represents the stop-gradient operator, which has partial derivatives equal to zero and remains constant during forward computation; $vq(x)$ denotes the output of Vector Quantization.

\textbf{Optimization.} Formally, our total loss function is defined as:
\begin{equation}
    \mathbb{L} = \mathbb{L}_{pred} + \lambda_1 \mathbb{L}_{ct}, %+ \lambda_2 \mathbb{L}_{Orth}
\end{equation}
where $\lambda_1>0$ is the hyperparameter.

%% file: tables_vq/different_vq_strcture_concise.tex
%\resizebox{!}{\.5\paperheight}{
% \vskip -0.2in
\begin{table*}[t]
\centering

%\begin{footnotesize}
% \begin{adjustwidth}{-1.5in}{-1in}
\caption{Univariate long-term series forecasting results of different sturcture of VQ. The best results are in bold. A lower MSE indicates better performance.The results are
averaged from 4 different prediction lengths $\in \{96, 192, 336, 720\}$. Appendix~\ref{appendix_various_vq} shows the full results.}\vspace{-1mm}

\begin{center}
% \begin{small}
\begin{sc}
\scalebox{0.80}{
\begin{tabular}{c|c|cccccccccccccccc}
\toprule
% \multicolumn{2}{c|}{Methods}&\multicolumn{2}{c|}{PatchTST(96)}&\multicolumn{2}{c|}{Attention}&\multicolumn{2}{c|}{Series-LG/GL}&\multicolumn{2}{c|}{Series-GL}&\multicolumn{2}{c}{Concatenate}\\
Methods&\multicolumn{2}{c|}{$Sparse-VQ$}&\multicolumn{2}{c|}{$VQ$}&\multicolumn{2}{c|}{$VQ_{cosine}$}&\multicolumn{2}{c|}{$VQ_{kmeans}$}&\multicolumn{2}{c|}{$VQ_{recursive}$}&\multicolumn{2}{c}{$VQ_{AdaptiveCodebook}$}\\
\midrule
Metric & MSE  & MAE & MSE & MAE& MSE  & MAE& MSE  & MAE& MSE  & MAE & MSE  & MAE\\
\midrule
ECL & \textbf{0.245} & \textbf{0.348} & 0.257  & 0.355   & 0.265  & 0.365  & 0.263           & 0.358      & 0.258   & 0.357    & 0.260          & 0.358      \\

\midrule
Traffic& \textbf{0.117} & \textbf{0.190} & 0.118  & 0.193   & 0.123  & 0.202  & 0.120           & 0.193      & 0.122   & 0.199     & \textbf{0.117} & 0.191      \\

\midrule
Weather& 0.0013         & \textbf{0.025} & 0.0013 & 0.0252   & 0.0013 & 0.0259 & \textbf{0.0012} & \textbf{0.025} & 0.0012 & 0.025   & 0.0013    & \textbf{0.025}   \\

\midrule
\bottomrule
\end{tabular}
\label{tab:different_vq_structure_concise}
}

\end{sc}
% \end{small}
\end{center}
%\vskip -0.15in
% \end{adjustwidth}
%\end{footnotesize}
\end{table*}
%}

%% file: tables_vq/EncoderVQ.tex
%\resizebox{!}{\.5\paperheight}{
\vskip -0.1in
\begin{table}[h]
\centering

%\begin{footnotesize}
% \begin{adjustwidth}{-1.5in}{-1in}
\caption{Univariate long-term series forecasting results of traditional transformer with Vector Quantization incorporated before the encoder(VQ-pre) and after the decoder(VQ-post). Input length $=512$ and prediction length $=96$. A lower MSE indicates better performance. All experiments are repeated three times.}\vspace{-3mm}

\begin{center}
% \begin{small}
\begin{sc}
\scalebox{0.9}{
\begin{tabular}{c|cccccccccccccc}
\toprule
% \multicolumn{2}{c|}{Methods}&\multicolumn{2}{c|}{PatchTST(96)}&\multicolumn{2}{c|}{Attention}&\multicolumn{2}{c|}{Series-LG/GL}&\multicolumn{2}{c|}{Series-GL}&\multicolumn{2}{c}{Concatenate}\\
Methods &\multicolumn{2}{c|}{PatchTST} &\multicolumn{2}{c|}{VQ-pre}&\multicolumn{2}{c}{VQ-post}\\
\midrule
Metric & MSE  & MAE & MSE & MAE  & MSE & MAE  \\
\midrule

 ettm2    & \textbf{0.065}  & \textbf{0.187}  & 0.067  & 0.191  &0.068 &0.196 \\
 electricity      & \textbf{0.209}  & \textbf{0.321}  & 0.226  & 0.331  &0.237 &0.341 \\
 traffic   & 0.134  & 0.223  & \textbf{0.125}  & \textbf{0.203}  &0.135 &0.219 \\
 weather   & 0.0013 & 0.0265 & \textbf{0.0009} & \textbf{0.0217}  & \textbf{0.0009} &0.0223\\
 wind     & \textbf{2.563}  & \textbf{1.261}  & 2.875  & 1.295  &3.017 &1.322 \\
 nordpool  & \textbf{0.856}  & 0.714  & 0.912  & \textbf{0.712}  &0.950 &0.719 \\
 caiso     & 0.162  &0.279  & \textbf{0.161}  & \textbf{0.277}  &0.169 &0.284 \\

\midrule
\bottomrule
\end{tabular}
\label{tab:ablation_EncoderVQ}
}

\end{sc}
% \end{small}
\end{center}
%\vskip -0.2in
% \end{adjustwidth}
%\end{footnotesize}
\end{table}
%}

%% file: tables_vq/PatchTST_woReverseNorm.tex
%\resizebox{!}{\.5\paperheight}{
% \vskip -0.2in
\begin{table*}[t]
\centering
\begin{sc}
%\color{blue}
\caption{Univariate forecasting results by PatchTST without ReverseNorm.}
\vspace{-1mm}
\scalebox{0.75}{

\begin{tabular}{c|c|cccccccccccccccccccc}

\toprule
\multicolumn{2}{c|}{woReverseNorm}
&\multicolumn{4}{c|}{traffic}&\multicolumn{4}{c|}{weather}&\multicolumn{4}{c}{ETTm2}\\
\midrule

\multicolumn{2}{c|}{PredLen} & 96 & 192  & 336 &720  & 96 & 192  & 336 & 720 & 96 & 192  & 336 & 720 \\

\midrule

\multirow{2}{*}{PatchTST\_wFFN} 
&MSE &\textbf{0.393} & \textbf{0.403} & \textbf{0.414} & \textbf{0.404} & 0.0450 & \textbf{0.0418} & \textbf{0.0347} & \textbf{0.0371} & \textbf{0.435} & \textbf{0.417} & \textbf{0.487} & \textbf{0.439} \\
&MAE &\textbf{0.454} & \textbf{0.460} & \textbf{0.468} & \textbf{0.462} & 0.1643 & \textbf{0.1622} & \textbf{0.1469} & \textbf{0.1517} & \textbf{0.455} & \textbf{0.433} & \textbf{0.464} & \textbf{0.425} \\

\midrule

\multirow{2}{*}{PatchTST\_woFFN}
&MSE &0.425          & 0.415          & 0.416          & 0.416          & \textbf{0.0442}          & 0.0463          & 0.0364          & 0.0376          & 0.569          & 0.619          & 0.602          & 0.591          \\
&MAE &0.479          & 0.470          & 0.470          & 0.467          & \textbf{0.1640}          & 0.1706          & 0.1505          & 0.1540          & 0.558          & 0.603          & 0.600          & 0.585      \\

\midrule
% \multicolumn{2}{c|}{Average} & & & & & & & & & & & & & & & \\
\bottomrule

\end{tabular}

}

\label{tab:PatchTST_woReverseNorm}
% \end{adjustwidth}
%\end{footnotesize}
\end{sc}
% \end{small}
\vskip -0.1in
\end{table*}
%}

%% file: tables_vq/parameters.tex
%\resizebox{!}{\.5\paperheight}{
% \vskip -0.2in
\begin{table}[t]
\begin{center}
% \begin{small}
\begin{sc}

%\color{blue}
\caption{Model parameter comparison for PatchTST and PatchTST with FFN-free structure, both the input length $=512$, univariate prediction length $O\in \{96, 192, 336, 720\}$. The amount of parameters is expressed in millions (M).}
\vspace{-1mm}
\scalebox{0.85}{
\begin{tabular}{c|ccccccc}
\toprule
%I’m 
{$Param(M)$} &{96} &{192} &{336} &{720} \\
\midrule

PatchTST  &0.604	&0.801	&1.096	&1.883	\\
$FFN_{free}$  &0.406	&0.603	&0.898	&1.685 \\
Reduction  & $\downarrow32.78\%$ & $\downarrow 24.72\%$ & $\downarrow 18.07\%$  & $\downarrow 10.52\%$ \\

\midrule
%average21.52%
\bottomrule
\end{tabular}
}
\label{tab:FFN_parameters}
% \end{adjustwidth}
%\end{footnotesize}
\end{sc}
% \end{small}
\end{center}
\vskip -0.1in
\end{table}
%}

%% file: sections/4_experiments.tex
\section{Experiments}
\label{sec_experiments}
%{\color{blue}

\subsection{Dataset and implementation details}
We have rigorously assessed our proposed Sparse-VQ model across ten well-established real-world benchmarks, encompassing ETT (m1, m2, h1, h2), Electricity, Traffic, Weather, Wind, Nordpool, and Caiso. For brevity, only the results for the ETTm2 dataset are presented. Comprehensive details about the datasets and the nuances of our implementation are available in Appendix \ref{sec_supplemental_experiments}.

\input{tables_vq/full_bench_uni_concise}

\input{tables_vq/full_bench_multi}

%\input{tables_vq/full_bench_multi}

%\paragraph{Baselines} We select representative baselines and cite their results from FPT Zhou et al. (2023)~\cite{zhou2023onefitsall}, which includes the most recent and quite extensive empirical studies of time series. The baselines include CNN-based models: TimesNet Wu et al. (2023)~\cite{TIMESNET} and Dlinear  Zeng et al. (2023)~\cite{Dlinear}; Transformer-based models: Informer Zhou et al. (2021)~\cite{haoyietal-informer-2021}, Autoformer Wu et al. (2021) \cite{Autoformer}, FEDformer Zhou et al. (2022) \cite{FedFormer}, Nonstationary Transformer Liu et al. (2022) \cite{Non-stationary-Transformers}, PatchTST Nie et al. (2022) \cite{patchTST}.Besides, N-BEATS Oreshkin et al. (2019) \cite{nbeats} are used for short-term forecasting.

\subsection{Long-term Forecasting}
For better comparison, we follow the experiment settings of Patch\-TST~\cite{patchTST}  where the input length is fixed to 512, and the prediction lengths for both training and evaluation are fixed to be 96, 192, 336, and 720, respectively.

\paragraph{Univariate Results} For univariate time series
forecasting, Sparse-VQ achieves the best performance on all seven benchmark datasets at all horizons as shown in Table~\ref{tab:full_bench_uni_concise}. Compared to PatchTST, Sparse-VQ yields an overall 7.84\% relative MAE reduction. On some datasets, such as traffic and weather, the improvement is more than 15\%. The experimental results in Table~\ref{tab:full_bench_uni_concise} verifies that Sparse-VQ is very effective in long-term forecasting.

\paragraph{Multivariate Results} The results for multivariate time series forecasting are summarized in Table~\ref{tab:full_bench_multi}. Compared with PatchTST, the proposed Sparse-VQ yields an overall 4.17\% relative MAE reduction. Overall, the improvement made by Sparse-VQ is consistent with varying horizons, implying its strength in long term forecasting.

\input{tables_vq/short_forecasting_average}

\input{tables_vq/few_shot_5_average}

\subsection{Short-term Forecasting}
To thoroughly assess different algorithms in forecasting, we extend our experiments to the M4 dataset~\cite{Makridakis2018TheMC} for short-term forecasting, which consists of univariate marketing data across yearly, quarterly, and monthly frequencies with a comparatively brief forecast horizon. Unlike the long-term datasets that feature a single continuous series from which samples are drawn using sliding windows, the M4 dataset is composed of 100,000 unique series collected at varying intervals.

Table \ref{tab:short_forecasting_average} highlights Sparse-VQ's superior performance against both advanced transformer-based and MLP-based models. Notably, when pitted against the similar transformer-based PatchTST method, Sparse-VQ achieves improvements of 1\%, 1.8\%, and 1.4\% in SMAPE, MASE, and OWA, respectively, and it exhibits performance on par with the N-BEATS method.

\subsection{Few-shot Forecasting}

Few-shot learning poses a unique challenge in forecasting, where models must forge robust representations from scant data. To probe Sparse-VQ's capability for time series analysis under this stringent condition, we devised targeted experiments. Deviating from the usual division of data into training, validation, and test segments, our few-shot approach operates with a mere sliver (e.g., 5\%, 10\%) of the training data. Table~\ref{tab:few_shot_5_average} encapsulates Sparse-VQ's impressive average outcomes using only 5\% of the data, outstripping recent cutting-edge methods such as OFA and PatchTST.Remarkably, in comparison to the transformer-based PatchTST method, Sparse-VQ records average enhancements of \textbf{9.5}\% in mse and \textbf{7.2}\% in mae across all benchmarks. A full exposition of our extensive findings over ten datasets is available in Appendix~\ref{subsec_appendix_few_shot}.

\subsection{Ablations}

\subsubsection{Validity analysis of several modules}

The ablation study in Table~\ref{tab:ablation_modules_average} provides important insights into how the Sparse-VQ and FFN-free structure affect the performance of our framework. We compare our results with PatchTST, which is considered the state-of-the-art benchmark for transformer-based models. By analyzing the results with and without Sparse-VQ / FFN-free structure, we can observe that both of these factors play significant roles in improving the forecasting performance.

When using the FFN-free structure, we observe a \textbf{4.36\%} decrease in MAE compared to the original model, suggesting that the FFN-free structure contributes to enhancing the model's forecasting performance. Furthermore, building upon the the FFN-free structure, integration of Sparse-VQ can achieve  a \textbf{5.04\%} reduction in MAE, whereas incorporating VQ yields a mere 1.21\% decrease by comparison, highlighting the importance of the sparse structure.

\begin{figure*}[h]
\centering
\setlength{\abovecaptionskip}{0.2cm}
% \scalebox{0.70}{
\includegraphics[width=1\linewidth, trim=0 0 0 0,clip]{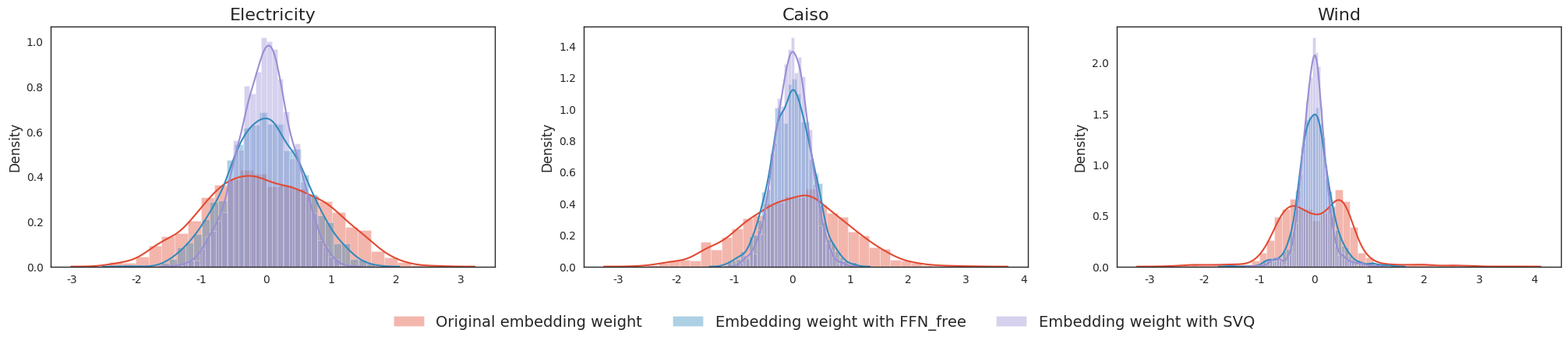}
\caption{Distribution of embedding weight. Sparse-VQ framework encourages denser weights}
\label{fig:sparse}
\vskip -0.1in
\end{figure*}

\begin{figure}[h]
\centering
\setlength{\abovecaptionskip}{0.2cm}
\vskip -0.08in
\scalebox{0.95}{
\includegraphics[width=1\linewidth, trim=70 10 70 10,clip]{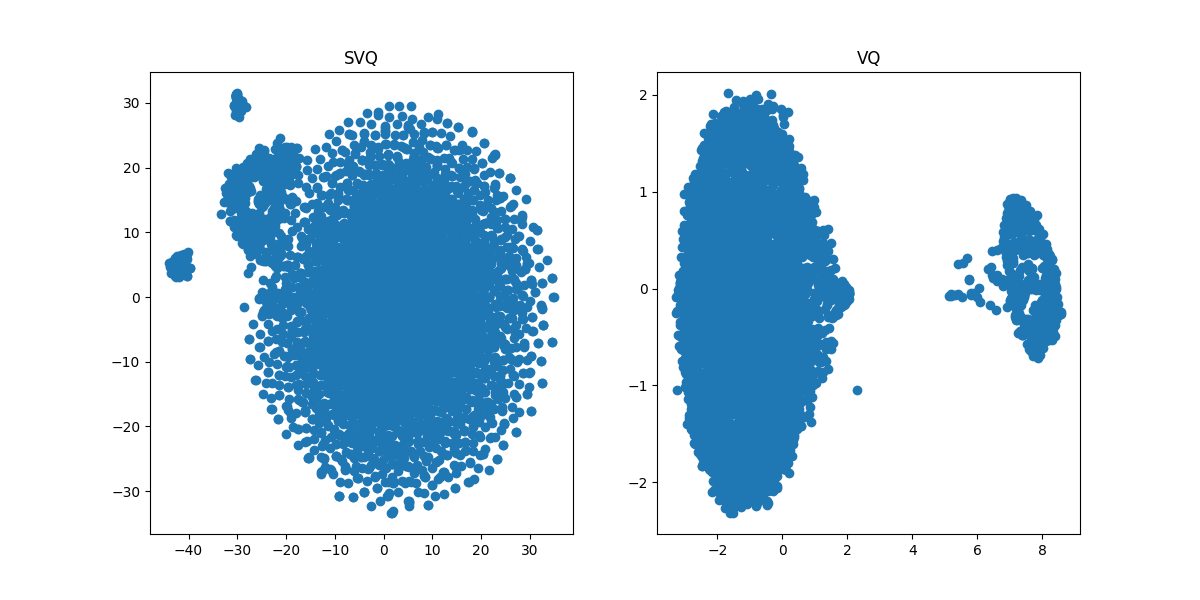}
}
\caption{ Distribution of codebook. Sparse-VQ encourage a sparser codebook with a wider range of perception.}
\label{fig:codebook}
\vskip -0.05in
\end{figure}

To explore how Sparse-VQ and FFN-free structures enhance the predictive accuracy of the model, we generate and plot the distribution of the model embeddings. Figure~\ref{fig:sparse} shows that both the two modules effectively concentrate the distribution of embeddings, thereby improving the performance of the model. Also, we employed t-SNE to reduce the dimensions of the codebooks from VQ and Sparse-VQ, facilitating their visualization as depicted in the Figure~\ref{fig:codebook}. The results clearly demonstrate that Sparse-VQ encompasses a wider representational range and presents a distribution that is more uniform and continuous than VQ.

\input{tables_vq/ablation_modules_average}

\subsubsection{Sparse-VQ}
The codebook size dictates the embedding capacity for diverse time series patterns and consequently affects sparse-VQ's performance. As evidenced in Table~\ref{tab:ablation_codebook_size}, enlarging the codebook size generally enhances model accuracy. However, excessively large codebooks can hinder convergence and ultimately reduce accuracy. 

Moreover, we would like to highlight the generality of our Sparse-VQ, which can serve as a plug-in to enhance the performance of other models. To demonstrate the general applicability of Sparse-VQ, we employ it in FEDformer~\cite{FedFormer}, Autoformer~\cite{Autoformer}. The results are summarized in Table \ref{tab:ablation_vq_boosting_average}. Integrating the Vector Quantization structure yields a modest boost, enhancing the FEDformer by 2.01\% and the Autoformer by 3.63\% in MSE. Notably, these improvements are consistent across models with varying predictive capabilities, suggesting that this could be a beneficial addition following the development of base models.

\input{tables_vq/ablation_vq_boosting_average}

\subsubsection{Robustness analysis}

\input{tables_vq/ablation_noise_VQ_average}

\input{tables_vq/ablation_codebook_size}

To evaluate the robustness of our model, we follow a commonly used practice by introducing noise into the data and then training the modified dataset, which is based on the approach described in the MICN~\cite{wang2023micn}. The experimental results are presented in Table~\ref{tab:ablation_noise_VQ_average}. The results show a slight increase in both metrics of Sparse-VQ as the amount of injected noise increases, while the prediction effect of other models have large fluctuations. Owing to space constraints, we present comprehensive results in Appendix~\ref{subsec_appendix_robustness}. For instance, under 10\% noise, Sparse-VQ exhibits a relative MAE increase of 6.3\%, compared to 8.9\% for PatchTST and 12.7\% for FEDformer, as detailed in appendix Tables ~\ref{tab:appendix_noise_PatchTST} and ~\ref{tab:appendix_noise_FEDformer}. Our findings reveal that our model, bolstered by its unique quantization approach, surpasses baseline counterparts in noise robustness.

%% file: tables_vq/full_bench_uni_concise.tex
\begin{table*}[h]
\centering
\begin{sc}

% \begin{adjustwidth}{-1.5in}{-1in}
\caption{Univariate long-term forecasting task. All the results are averaged from 4 different prediction lengths $\in \{96,192,336,720\}$. A lower MAE indicates better performance. All experiments are repeated 3 times. Appendix \ref{subsec_appendix_long_forecasting}  shows the full results.}\vspace{-3mm}

%Univariate long-term series forecasting results with same input length $=512$ and prediction length $\in \{96,192,336,720\}$. A lower MAE indicates better performance. All experiments are repeated 3 times. Full results are in appendix .}\vspace{-1mm}
\scalebox{0.75}{
\begin{tabular}{cc|ccccccccccccccccccc}

\toprule
&{Methods} &\multicolumn{2}{c|}{SVQ} &\multicolumn{2}{c|}{PatchTST} &\multicolumn{2}{c|}{OFA}  &\multicolumn{2}{c|}{Dlinear}  &\multicolumn{2}{c|}{FEDformer} &\multicolumn{2}{c|}{Autoformer} &\multicolumn{2}{c|}{Informer} &\multicolumn{2}{c|}{LogTrans} &\multicolumn{2}{c}{Reformer}\\
\midrule
&{Metric} & MSE  & MAE & MSE & MAE & MSE  & MAE & MSE  & MAE& MSE  & MAE& MSE  & MAE& MSE  & MAE & MSE  & MAE & MSE & MAE\\
\midrule
&{ETTm2}

& \textbf{0.111}  & \textbf{0.248}  & 0.113  & 0.252  & 0.119  & 0.272  & 0.112  & \textbf{0.248} & 0.118  & 0.2595 & 0.130  & 0.271 & 0.175  & 0.320 & 0.130  & 0.277 & 0.134  & 0.288 \\

\midrule
&{Electricity}
& \textbf{0.245}  & \textbf{0.348}  & 0.267  & 0.367  & 0.292  & 0.380  & 0.252  & 0.355          & 0.326  & 0.418  & 0.414  & 0.479 & 0.624  & 0.598 & 0.410  & 0.473 & 0.352  & 0.435 \\

\midrule
&{Traffic}
 & \textbf{0.117}  & \textbf{0.190}  & 0.136  & 0.228  & 0.156  & 0.261  & 0.128  & 0.204          & 0.177  & 0.27   & 0.261  & 0.365 & 0.309  & 0.388 & 0.355  & 0.404 & 0.375  & 0.434 \\

\midrule
&{Weather}
 & \textbf{0.0013} & \textbf{0.0253} & 0.0016 & 0.0294 & 0.0017 & 0.0311 & 0.0061 & 0.0645         & 0.007  & 0.0615 & 0.0083 & 0.07  & 0.0033 & 0.044 & 0.0058 & 0.057 & 0.0077 & 0.069 \\

\midrule
&{Wind}

 & \textbf{2.794}  & \textbf{1.351}  & 3.084  & 1.414  & 3.456  & 1.530  & 3.207  & 1.425          & 3.808  & 1.638  & 3.821  & 1.648 & 3.960  & 1.506 & 4.654  & 1.622 & 3.851  & 1.489 \\

\midrule
&{Nordpool}
 & \textbf{0.798}  & \textbf{0.684}  & 0.890  & 0.740  & 0.896  & 0.743  & 0.887  & 0.743          & 0.876  & 0.754  & 0.999  & 0.8   & 0.849  & 0.728 & 0.882  & 0.729 & 0.873  & 0.739 \\

\midrule
&{Caiso}
 & \textbf{0.233}  & \textbf{0.324}  & 0.241  & 0.340  & 0.264  & 0.3495 & 0.243  & 0.331          & 0.269  & 0.376  & 0.323  & 0.414 & 0.299  & 0.4   & 0.288  & 0.393 & 0.271  & 0.380\\

\midrule
% \multicolumn{2}{c|}{Average} & & & & & & & & & & & & & & & \\
\bottomrule
\end{tabular}
\label{tab:full_bench_uni_concise}
}
% \end{adjustwidth}
%\vskip -0.1in
\vskip -0.1in
\end{sc}
\end{table*}
%}

%% file: tables_vq/full_bench_multi.tex
%\resizebox{!}{\.5\paperheight}{
% \vskip -0.2in
\begin{table*}[h]
\centering
\begin{sc}

%\begin{adjustwidth}{-1.5in}{-1in}
\caption{Multivariate long-term series forecasting results on four datasets with same input length $=512$ and various prediction lengths $\in \{96,192,336,720\}$ . A lower MAE indicates better performance. All experiments are repeated 3 times. Full results of four ETT datasets are in Appendix \ref{subsec_appendix_long_forecasting} .}\vspace{-3mm}
\scalebox{0.85}{
\begin{tabular}{c|c|cccccccccccccccccc}

\toprule
\multicolumn{2}{c|}{Methods} &\multicolumn{2}{c|}{SVQ} &\multicolumn{2}{c|}{PatchTST} &\multicolumn{2}{c|}{OFA}  &\multicolumn{2}{c|}{Dlinear}  &\multicolumn{2}{c|}{FEDformer} &\multicolumn{2}{c|}{Autoformer} &\multicolumn{2}{c|}{Informer} &\multicolumn{2}{c|}{LogTrans} &\multicolumn{2}{c}{Reformer}\\
\midrule
\multicolumn{2}{c|}{Metric} & MSE  & MAE & MSE & MAE & MSE  & MAE & MSE  & MAE& MSE  & MAE& MSE  & MAE& MSE  & MAE & MSE  & MAE & MSE & MAE\\

\midrule
\multirow{5}{*}{\rotatebox{90}{ETTm2}} 

& 96  & 0.158 & 0.242 & 0.166 & 0.256 & 0.173 & 0.262 & 0.167 & 0.26  & 0.203 & 0.287 & 0.255 & 0.339 & 0.705 & 0.69  & 0.768 & 0.642 & 0.365 & 0.453 \\
 & 192 & 0.215 & 0.282 & 0.223 & 0.296 & 0.229 & 0.301 & 0.224 & 0.303 & 0.269 & 0.328 & 0.281 & 0.34  & 0.924 & 0.692 & 0.989 & 0.757 & 0.533 & 0.563 \\
 & 336 & 0.268 & 0.317 & 0.274 & 0.329 & 0.286 & 0.341 & 0.281 & 0.342 & 0.325 & 0.366 & 0.339 & 0.372 & 1.364 & 0.877 & 1.334 & 0.872 & 1.363 & 0.887 \\
 & 720 & 0.349 & 0.371 & 0.362 & 0.385 & 0.378 & 0.401 & 0.397 & 0.421 & 0.421 & 0.415 & 0.422 & 0.419 & 0.877 & 1.074 & 3.048 & 1.328 & 3.379 & 1.338 \\
 & Avg &\textbf{ 0.248} & \textbf{0.303} & 0.256 & 0.317 & 0.267 & 0.326 & 0.267 & 0.332 & 0.305 & 0.349 & 0.324 & 0.368 & 0.968 & 0.833 & 1.535 & 0.900 & 1.41  & 0.810 \\

\midrule
\multirow{5}{*}{\rotatebox{90}{Electricity}}
 & 96  & 0.127 & 0.216 & 0.129 & 0.222 & 0.139 & 0.238 & 0.14  & 0.237 & 0.183 & 0.297 & 0.201 & 0.317 & 0.304 & 0.405 & 0.258 & 0.357 & 0.274 & 0.368 \\
 & 192 & 0.144 & 0.233 & 0.147 & 0.24  & 0.153 & 0.251 & 0.153 & 0.249 & 0.195 & 0.308 & 0.222 & 0.334 & 0.313 & 0.413 & 0.266 & 0.368 & 0.296 & 0.386 \\
 & 336 & 0.161 & 0.251 & 0.163 & 0.259 & 0.169 & 0.266 & 0.169 & 0.267 & 0.212 & 0.313 & 0.231 & 0.338 & 0.29  & 0.381 & 0.28  & 0.38  & 0.3   & 0.394 \\
 & 720 & 0.197 & 0.284 & 0.197 & 0.29  & 0.206 & 0.297 & 0.203 & 0.301 & 0.231 & 0.343 & 0.254 & 0.361 & 0.262 & 0.344 & 0.283 & 0.376 & 0.373 & 0.439 \\
 & Avg &\textbf{ 0.157} & \textbf{0.246} & 0.159 & 0.253 & 0.167 & 0.263 & 0.166 & 0.264 & 0.205 & 0.315 & 0.227 & 0.338 & 0.292 & 0.386 & 0.272 & 0.370 & 0.311 & 0.397 \\

\midrule
\multirow{5}{*}{\rotatebox{90}{Traffic}}
 & 96  & 0.377 & 0.241 & 0.36  & 0.249 & 0.388 & 0.282 & 0.41  & 0.282 & 0.562 & 0.349 & 0.613 & 0.388 & 0.824 & 0.514 & 0.684 & 0.384 & 0.719 & 0.391 \\
 & 192 & 0.390 & 0.247 & 0.379 & 0.256 & 0.407 & 0.290 & 0.423 & 0.287 & 0.562 & 0.346 & 0.616 & 0.382 & 1.106 & 0.672 & 0.685 & 0.39  & 0.696 & 0.379 \\
 & 336 & 0.399 & 0.252 & 0.392 & 0.264 & 0.412 & 0.294 & 0.436 & 0.296 & 0.57  & 0.323 & 0.622 & 0.337 & 1.084 & 0.627 & 0.733 & 0.408 & 0.777 & 0.42  \\
 & 720 & 0.438 & 0.275 & 0.432 & 0.286 & 0.450 & 0.312 & 0.466 & 0.315 & 0.596 & 0.368 & 0.66  & 0.408 & 1.536 & 0.845 & 0.717 & 0.396 & 0.864 & 0.472 \\
 & Avg & 0.401 & \textbf{0.254} & \textbf{0.391} & 0.264 & 0.414 & 0.295 & 0.434 & 0.295 & 0.573 & 0.347 & 0.628 & 0.379 & 1.138 & 0.665 & 0.705 & 0.395 & 0.764 & 0.416 \\

\midrule
\multirow{5}{*}{\rotatebox{90}{Weather}}
 & 96  & 0.145 & 0.183 & 0.149 & 0.198 & 0.162 & 0.212 & 0.176 & 0.237 & 0.217 & 0.296 & 0.266 & 0.336 & 0.406 & 0.444 & 0.458 & 0.49  & 0.3   & 0.384 \\
 & 192 & 0.188 & 0.228 & 0.194 & 0.241 & 0.204 & 0.248 & 0.22  & 0.282 & 0.276 & 0.336 & 0.307 & 0.367 & 0.525 & 0.527 & 0.658 & 0.589 & 0.598 & 0.544 \\
 & 336 & 0.238 & 0.269 & 0.245 & 0.282 & 0.254 & 0.286 & 0.265 & 0.319 & 0.339 & 0.38  & 0.359 & 0.395 & 0.531 & 0.539 & 0.797 & 0.652 & 0.578 & 0.523 \\
 & 720 & 0.306 & 0.321 & 0.314 & 0.334 & 0.326 & 0.337 & 0.323 & 0.362 & 0.403 & 0.428 & 0.578 & 0.578 & 0.419 & 0.428 & 0.869 & 0.675 & 1.059 & 0.741 \\
 & Avg &\textbf{0.219} & \textbf{0.250} & 0.226 & 0.264 & 0.237 & 0.271 & 0.246 & 0.3   & 0.309 & 0.36  & 0.378 & 0.419 & 0.470 & 0.485 & 0.696 & 0.602 & 0.634 & 0.548 \\

\midrule
\multirow{5}{*}{\rotatebox{90}{Wind}}
 & 96  & 0.886 & 0.622 & 0.894 & 0.639 & 0.942 & 0.657 & 0.902 & 0.649 & 1.438 & 0.892 & 1.475 & 0.900 & 1.422 & 0.803 & 1.497 & 0.849 & 1.097 & 0.705 \\
 & 192 & 1.133 & 0.748 & 1.166 & 0.782 & 1.168 & 0.774 & 1.124 & 0.766 & 1.572 & 0.945 & 1.720 & 0.973 & 1.810 & 0.950 & 1.574 & 0.911 & 1.279 & 0.783 \\
 & 336 & 1.275 & 0.841 & 1.374 & 0.875 & 1.387 & 0.872 & 1.329 & 0.862 & 1.815 & 1.031 & 1.667 & 0.968 & 1.714 & 0.926 & 1.551 & 0.918 & 1.471 & 0.897 \\
 & 720 & 1.418 & 0.910 & 1.545 & 0.955 & 1.587 & 0.955 & 1.495 & 0.935 & 1.788 & 1.021 & 1.738 & 1.000 & 1.877 & 0.990 & 1.602 & 0.952 & 1.502 & 0.944 \\
 & Avg &\textbf{ 1.178} & \textbf{0.780} & 1.245 & 0.813 & 1.271 & 0.815 & 1.213 & 0.803 & 1.653 & 0.972 & 1.65  & 0.960 & 1.706 & 0.917 & 1.556 & 0.908 & 1.337 & 0.832 \\

\midrule
\multirow{5}{*}{\rotatebox{90}{Nordpool}}
 & 96  & 0.543 & 0.540 & 0.560 & 0.558 & 0.563 & 0.556 & 0.613 & 0.593 & 0.552 & 0.579 & 0.815 & 0.702 & 1.029 & 0.788 & 0.858 & 0.691 & 0.645 & 0.618 \\
 & 192 & 0.608 & 0.579 & 0.632 & 0.594 & 0.613 & 0.590 & 0.674 & 0.636 & 0.646 & 0.634 & 0.795 & 0.697 & 1.030 & 0.804 & 0.918 & 0.715 & 0.728 & 0.659 \\
 & 336 & 0.593 & 0.578 & 0.596 & 0.594 & 0.595 & 0.587 & 0.653 & 0.628 & 0.599 & 0.593 & 0.829 & 0.713 & 1.198 & 0.860 & 0.953 & 0.744 & 0.782 & 0.675 \\
 & 720 & 0.585 & 0.580 & 0.586 & 0.590 & 0.580 & 0.587 & 0.642 & 0.627 & 0.634 & 0.624 & 0.736 & 0.672 & 1.246 & 0.865 & 0.997 & 0.759 & 0.836 & 0.694 \\
 & Avg &\textbf{ 0.582} & \textbf{0.569} & 0.594 & 0.584 & 0.588 & 0.58  & 0.646 & 0.621 & 0.608 & 0.608 & 0.794 & 0.696 & 1.126 & 0.829 & 0.932 & 0.727 & 0.748 & 0.662 \\

\midrule
\multirow{5}{*}{\rotatebox{90}{Caiso}} 
 & 96  & 0.204 & 0.286 & 0.237 & 0.307 & 0.210 & 0.297 & 0.221 & 0.302 & 0.265 & 0.365 & 0.327 & 0.406 & 0.320 & 0.394 & 0.261 & 0.355 & 0.242 & 0.343 \\
 & 192 & 0.279 & 0.341 & 0.323 & 0.364 & 0.276 & 0.349 & 0.279 & 0.350 & 0.325 & 0.413 & 0.462 & 0.499 & 0.415 & 0.457 & 0.295 & 0.378 & 0.285 & 0.372 \\
 & 336 & 0.331 & 0.376 & 0.370 & 0.402 & 0.330 & 0.389 & 0.328 & 0.388 & 0.345 & 0.421 & 0.584 & 0.552 & 0.473 & 0.492 & 0.368 & 0.458 & 0.343 & 0.455 \\
 & 720 & 0.430 & 0.428 & 0.456 & 0.458 & 0.466 & 0.456 & 0.450 & 0.459 & 0.416 & 0.471 & 0.505 & 0.517 & 0.536 & 0.533 & 0.506 & 0.503 & 0.498 & 0.489 \\
 & Avg &\textbf{ 0.311} & \textbf{0.357} & 0.347 & 0.383 & 0.321 & 0.373 & 0.320 & 0.375 & 0.338 & 0.418 & 0.470 & 0.494 & 0.436 & 0.469 & 0.358 & 0.424 & 0.342 & 0.415 \\

\midrule

% \midrule
% \multicolumn{2}{c|}{Average} & & & & & & & & & & & & & & & \\
\bottomrule
\end{tabular}
\label{tab:full_bench_multi}
}
% \end{adjustwidth}
%\vskip 0.1in
\vskip -0.1in
\end{sc}
\end{table*}
%}

%% file: tables_vq/short_forecasting_average.tex
%\resizebox{!}{\.5\paperheight}{
% \vskip -0.2in
\begin{table*}[t]
\centering
\begin{sc}

% \begin{adjustwidth}{-1.5in}{-1in}
\caption{Short-term forecasting task on M4. The prediction lengths are $\in \{6, 48\}$ and results are weighted
averaged from several datasets under different sample intervals. A lower score indicates better performance. All experiments are repeated 3 times.Full results are in Appendix \ref{subsec_appendix_short_forecasting}}\vspace{-1mm}
\scalebox{0.85}{
\begin{tabular}{c|c|cccccccccccccccccc}

\toprule

\multicolumn{2}{c|}{Methods} &{SVQ} &{OFA} &{PatchTST} &{N-HiTS} &{N-BEATS} &{ETSformer} &{LighTS} &{Dlinear} &{FEDformer} &{Autoformer} &{Informer} &{Reformer} \\
\midrule
\multirow{3}{*}{\rotatebox{90}{Average}} 

& SMAPE & 11.938 &{11.991} &12.059&11.927 &\textbf{11.851} &14.718 &13.525 &13.639 &12.840 &12.909 &14.086 &18.200 \\
& MASE & \textbf{1.593} & {1.600} &1.623 &1.613 &1.599 &2.408 &2.111 &2.095 &1.701 &1.771 &2.718 &4.223   \\
& OWA & 0.857 & {0.861} &0.869 &0.861 &\textbf{0.855} &1.172 &1.051 &1.051 &0.918&0.939 &1.230 &1.775 \\
\midrule

% \midrule
% \multicolumn{2}{c|}{Average} & & & & & & & & & & & & & & & \\
\bottomrule
\end{tabular}
\label{tab:short_forecasting_average}
}
% \end{adjustwidth}
%\vskip 0.1in
\vskip -0.1in
\end{sc}
\end{table*}
%}

%% file: tables_vq/few_shot_5_average.tex
%\resizebox{!}{\.5\paperheight}{
%\vskip -0.2in
\begin{table*}[h]
\centering
\begin{sc}

% \begin{adjustwidth}{-1.5in}{-1in}
\caption{Few-shot learning results on 5\% data. All the results are averaged from 4 different prediction lengths $\in \{96,192,336,720\}$ .A lower MSE indicates better performance, and the best results are highlighted in bold. Full results are in Appendix \ref{subsec_appendix_few_shot}}\vspace{-1mm}
\scalebox{0.85}{
\begin{tabular}{cc|cccccccccccccccccc}

\toprule
&{Methods} &\multicolumn{2}{c|}{SVQ} &\multicolumn{2}{c|}{PatchTST} &\multicolumn{2}{c|}{OFA}  &\multicolumn{2}{c|}{Dlinear}  &\multicolumn{2}{c|}{FEDformer} &\multicolumn{2}{c|}{Autoformer} &\multicolumn{2}{c|}{Informer} &\multicolumn{2}{c|}{LogTrans} &\multicolumn{2}{c}{Reformer}\\

\midrule
&{Metric} & MSE  & MAE & MSE & MAE & MSE  & MAE & MSE  & MAE& MSE  & MAE& MSE  & MAE& MSE  & MAE & MSE  & MAE & MSE & MAE\\
\midrule

&{ETTm1} & \textbf{0.405}  & \textbf{0.412}   & 0.526 & 0.476 & 0.472 & 0.45  & 0.400 & 0.417 & 0.73  & 0.592 & 0.796 & 0.62  & 1.163 & 0.791 & 1.597 & 0.979 & 1.264 & 0.826 \\

\midrule

&{Electricity} & 0.185 & 0.281 &0.181 & 0.277 & 0.178 & \textbf{0.273} & \textbf{0.176} & 0.275 & 0.266 & 0.353 & 0.346 & 0.404 & 1.281 & 0.929 & 0.934 & 0.746 & 1.289 & 0.904 \\

\midrule

&{Traffic} & 0.426  & \textbf{0.288}   & \textbf{0.418} & 0.296 & 0.434 & 0.305 & 0.45  & 0.317 & 0.676 & 0.423 & 0.833 & 0.502 & 1.591 & 0.832 & 1.309 & 0.685 & 1.618 & 0.851 \\

\midrule
&{Weather} & \textbf{0.258}  & \textbf{0.283}   & 0.269 & 0.303 & 0.263 & 0.301 & 0.263 & 0.308 & 0.309 & 0.353 & 0.31  & 0.353 & 0.584 & 0.527 & 0.457 & 0.458 & 0.447 & 0.453 \\

\midrule
&{Wind} & \textbf{1.321}  & \textbf{0.830}   & 1.469 & 0.892 & 1.489 & 0.898 & 1.396 & 0.869 & 1.741 & 1.005 & 1.893 & 1.038 & 4.183 & 1.622 & 3.847 & 1.558 & 2.594 & 1.320 \\

\midrule
&{Nordpool} & \textbf{0.654}  & \textbf{0.608}   & 0.711 & 0.642 & 0.752 & 0.656 & 0.71  & 0.646 & 0.954 & 0.766 & 0.994 & 0.784 & 3.158 & 1.430 & 2.171 & 1.188 & 1.99  & 1.142 \\

\midrule
&{Caiso} & \textbf{0.319}  & \textbf{0.366}   & 0.386 & 0.426 & 0.382 & 0.431 & 0.382 & 0.426 & 0.653 & 0.599 & 0.729 & 0.633 & 1.785 & 0.966 & 1.664 & 0.922 & 1.433 & 0.862\\

\midrule

%\multicolumn{2}{c|}{Average} & \textbf{0.316}  & \textbf{0.314} & 0.377 &0.375  &0.392 &0.383  &0.483 &0.445 &0.537 &0.480 &0.697 &0.537 &0.909 &0.675 &1.341 &0.789 &1.878 &0.994 &1.819 &0.966\\
% \midrule
% \multicolumn{2}{c|}{Average} & & & & & & & & & & & & & & & \\
\bottomrule
\end{tabular}
\label{tab:few_shot_5_average}
}
% \end{adjustwidth}
%\vskip 0.1in
\vskip -0.1in
\end{sc}
\end{table*}
%}

%% file: tables_vq/ablation_modules_average.tex
%\resizebox{!}{\.5\paperheight}{
%\vskip -0.05in
\begin{table}[h]
\centering

%\begin{footnotesize}
% \begin{adjustwidth}{-1.5in}{-1in}
\caption{Ablation study of FFN-free and Sparse-VQ in PatchTST. 4 cases are included: (a) both FFN-free and Sparse-VQ are included in model (SVQ+FFN-f); (b) only Vector Quantization (VQ+FFN-f); (c) only FFN-free(FFN-f);(d) neither of them is included (Original patchTST model). The best results are in bold. A lower MSE indicates better performance. Appendix~\ref{subsec_appendix_ablation} shows the full results. }\vspace{-3mm}

\begin{center}
% \begin{small}
\begin{sc}
\scalebox{0.75}{
\begin{tabular}{c|cccccccccccccccc}
\toprule
% \multicolumn{2}{c|}{Methods}&\multicolumn{2}{c|}{PatchTST(96)}&\multicolumn{2}{c|}{Attention}&\multicolumn{2}{c|}{Series-LG/GL}&\multicolumn{2}{c|}{Series-GL}&\multicolumn{2}{c}{Concatenate}\\
{Methods}&\multicolumn{2}{c|}{SVQ+FFN-f}&\multicolumn{2}{c|}{VQ+FFN-f}&\multicolumn{2}{c|}{FFN-f}&\multicolumn{2}{c}{Original}\\
\midrule
{Metric} & MSE  & MAE & MSE & MAE& MSE  & MAE& MSE  & MAE\\
\midrule
{ETTm2}
&\textbf{0.111} & \textbf{0.248} & \textbf{0.111} & 0.249 & 0.112 & 0.250 & 0.113 & 0.252 \\

\midrule
{Electricity}
&\textbf{0.245} & \textbf{0.348} & 0.253 & 0.353 & 0.256 & 0.355 & 0.267 & 0.365 \\

\midrule
{Weather}
&\textbf{0.0013} & \textbf{0.0253} & 0.0015 & 0.0280 & 0.0015 & 0.0287 & 0.0016 & 0.0294 \\

\midrule
{Traffic}
&\textbf{0.117} & \textbf{0.190} & 0.120 & 0.199 & 0.122 & 0.202 & 0.136 & 0.228 \\

\midrule
\bottomrule
\end{tabular}
\label{tab:ablation_modules_average}
}

\end{sc}
% \end{small}
\end{center}
\vskip -0.10in
% \end{adjustwidth}
%\end{footnotesize}
\end{table}
%}

%% file: tables_vq/ablation_vq_boosting_average.tex
%\resizebox{!}{\.5\paperheight}{

\begin{table}[h]
\vskip -0.05in
\centering
%\begin{footnotesize}
% \begin{adjustwidth}{-1.5in}{-1in}

\caption{Results for boosting effect of sparse-VQ.  All the results are averaged from 4 different prediction lengths $\in \{96, 192, 336, 720\}$. We use FEDformer and Autoformer as backbones and leverage them with the Sparse-VQ. A lower MSE indicates better performance. All experiments are repeated 3 times. Appendix~\ref{subsec_appendix_ablation} shows the full results.}
%\vspace{-2mm}
\begin{center}
\begin{small}
\begin{sc}

\scalebox{0.80}{
\begin{tabular}{c|ccccccccccccccc}
\toprule
{Methods}&\multicolumn{2}{c|}{FEDformer}&\multicolumn{2}{c|}{FEDformer+SVQ}&\multicolumn{2}{c|}{Autoformer}&\multicolumn{2}{c}{Autoformer+SVQ}\\
\midrule
{Metric} & MSE  & MAE & MSE & MAE& MSE  & MAE& MSE & MAE\\
\midrule
{ETTm2}
 &0.305 & \textbf{0.349} &\textbf{0.303} & 0.351 & 0.324 & 0.368 & \textbf{0.306} & \textbf{0.350} \\

\midrule
{Electricity}
 &0.214 & 0.327 & \textbf{0.209} & \textbf{0.323} & \textbf{0.227} & 0.338 & 0.232 & \textbf{0.333} \\

\midrule
{Traffic}
 &0.610 &0.376 & \textbf{0.606} & \textbf{0.373} & 0.628 & \textbf{0.379} & \textbf{0.625} & 0.389 \\

\midrule
{Weather}
 &0.309 & 0.360 & \textbf{0.296} & \textbf{0.348} & 0.338 & 0.382 & \textbf{0.308} & \textbf{0.356} \\
\midrule
\bottomrule
\end{tabular}
\label{tab:ablation_vq_boosting_average}
}
\end{sc}
\end{small}
\end{center}
\vskip -0.2in
% \end{adjustwidth}
%\end{footnotesize}
\end{table}
%}

%% file: tables_vq/ablation_noise_VQ_average.tex
%\resizebox{!}{\.5\paperheight}{

\begin{table}[h]
\vskip -0.02in
\centering
%\begin{footnotesize}
% \begin{adjustwidth}{-1.5in}{-1in}

\caption{Robustness analysis of univariate results conducted on four typical datasets. The degree of noise injected into the time series data is determined by $\eta$. All the results are
averaged from 4 different prediction lengths $\in \{96, 192, 336, 720\}$.All experiments are repeated 3 times. Appendix~\ref{subsec_appendix_robustness} shows the full results.}\vspace{-1mm}
\begin{center}
\begin{small}
\begin{sc}

\scalebox{0.85}{
\begin{tabular}{c|cccccccccccccccccc}
\toprule
{VQ}&\multicolumn{2}{c|}{Original}&\multicolumn{2}{c|}{$\eta=1\%$}&\multicolumn{2}{c|}{$\eta=5\%$}&\multicolumn{2}{c}{$\eta=10\%$}\\
\midrule
{Metric} & MSE  & MAE & MSE & MAE& MSE  & MAE& MSE & MAE\\
\midrule
{ETTm2}
&0.111	&0.248	&0.112	&0.250	&0.114	&0.253	&0.114	&0.255\\

\midrule
{Electricity}
&0.253	&0.353	&0.254	&0.355	&0.258	&0.356	&0.265	&0.368 \\

\midrule
{Traffic}
&0.120	&0.199	&0.121	&0.201	&0.125	&0.212	&0.138	&0.239 \\

\midrule
{Weather}
&0.0015	&0.028	&0.0015	&0.028	&0.0015	&0.0282	&0.0015	&0.029 \\

\midrule
\bottomrule

\end{tabular}

}
\label{tab:ablation_noise_VQ_average}
\end{sc}
\end{small}
\end{center}
\vskip -0.1in
% \end{adjustwidth}
%\end{footnotesize}
\end{table}
%}

%% file: tables_vq/ablation_codebook_size.tex
%\resizebox{!}{\.5\paperheight}{
\vskip -0.05in
\begin{table}[h]
\centering

%\begin{footnotesize}
% \begin{adjustwidth}{-1.5in}{-1in}
\caption{SVQ with different size of codebook. A lower MSE indicates better performance. All experiments are repeated three times on average.} %Input length $=512$ and prediction length $=720$ with hidden dimension $=16$. 
\vspace{-3mm}

\begin{center}
% \begin{small}
\begin{sc}
\scalebox{0.80}{
\begin{tabular}{cc|cccccccccc}
\toprule

&{Codebook Size} &\multicolumn{2}{c|}{10} &\multicolumn{2}{c|}{750}&\multicolumn{2}{c}{10000}\\
\midrule
&Metric & MSE  & MAE & MSE & MAE  & MSE & MAE \\
\midrule

&Caiso &0.500 & 0.471 & \textbf{0.451} & \textbf{0.454} & 0.480 &463  \\
\midrule
&Weather &0.315 & 0.341 & \textbf{0.306} & \textbf{0.322} & 0.322 & 0.349 \\
\midrule
&ETTm2 &0.372 & 0.393  & \textbf{0.356} & \textbf{0.376} & 0.367 & 0.389 \\

\midrule
\bottomrule
\end{tabular}
\label{tab:ablation_codebook_size}
}

\end{sc}
% \end{small}
\end{center}
\vskip -0.10in
% \end{adjustwidth}
%\end{footnotesize}
\end{table}
%}

%% file: sections/5_conclusions.tex
\section{Conclusion}
\label{sec_conclusion}

In summary, our research addresses the unique challenges of time series analysis—distribution shifts and high noise levels—by proposing the Sparse Vector Quantized FFN-Free Transformer (Sparse-VQ). This innovative model reimagines the transformer architecture sans the conventional Feed-Forward layer, utilizing Sparse-VQ and Reverse Instance Normalization (RevIN) for noise reduction and statistical capture. This leads to a more efficient model with fewer parameters, which not only reduces overfitting but also enhances computational performance. Our Sparse-VQ model has demonstrated superior accuracy, outperforming established models with significant reductions in MAE on average (7.84\% for univariate and 4.17\% for multivariate forecasting), as validated on ten benchmark datasets, including the novel CAISO dataset. Furthermore, Sparse-VQ's design allows for easy integration with existing transformer-based models, improving their efficacy in time series forecasting. We encourage the community to explore beyond the adaptation of traditional transformers for time series analysis by meticulously dissecting transformer components. A more efficient alternative tailored to these applications may well be within closer reach than anticipated.

%% file: sections/6_Appendix.tex
\input{appendix_vq/appendix_relatedworks}

\input{appendix_vq/appendix_supplemental_experiments}

\input{appendix_vq/appendix_proof}

\begin{figure*}[h]
\centering
\setlength{\abovecaptionskip}{0.2cm}
% \scalebox{0.70}{
\includegraphics[width=1\linewidth, trim=0 0 0 0,clip]{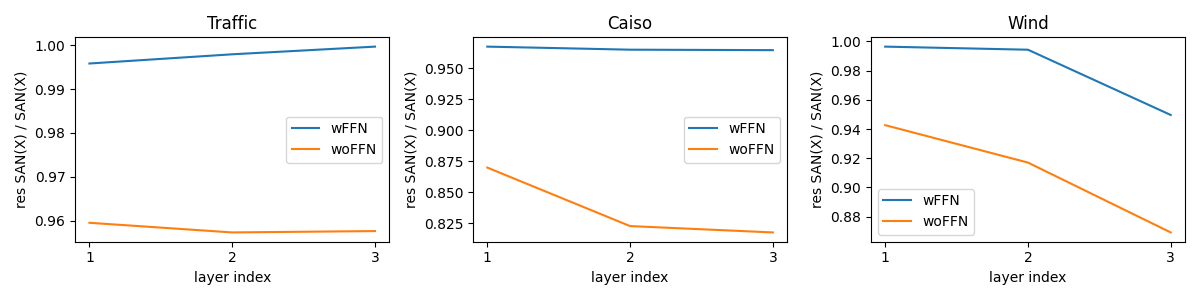}
\caption{Relative norm of the residual along the depth for PatchTST with and without FFN}
\label{fig:FFN_collapse}
% \vskip -0.2in
\end{figure*}

\begin{figure*}[h]
\centering
\setlength{\abovecaptionskip}{0.2cm}
\scalebox{0.90}{
\includegraphics[width=1\linewidth, trim=0 0 0 0,clip]{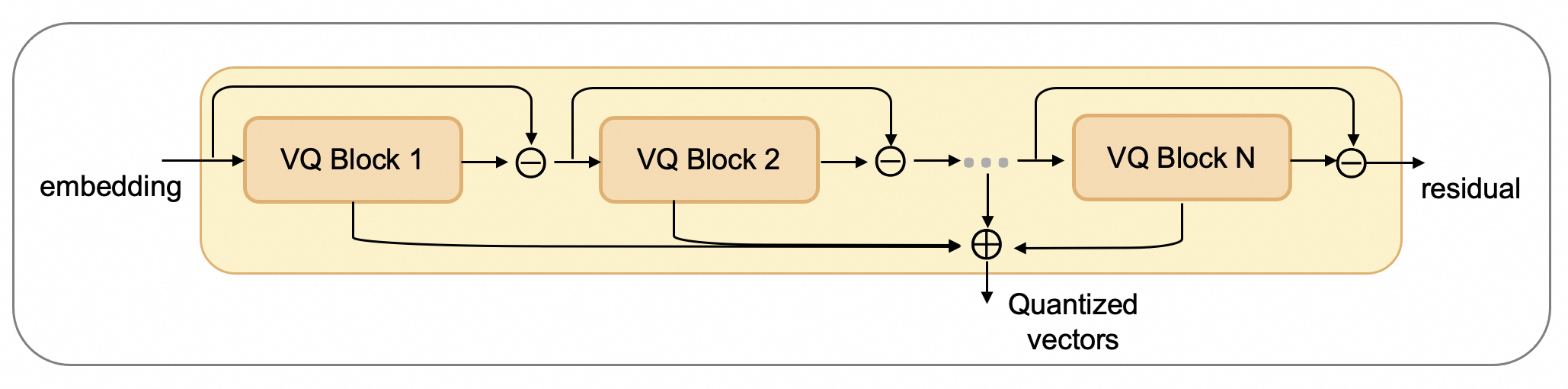}}
\caption{Recursive structure of VQ}
\label{fig:VQ_recursive}
% \vskip -0.2in
\end{figure*}

\begin{figure*}[h]
\centering
\setlength{\abovecaptionskip}{0.2cm}
\scalebox{0.90}{
\includegraphics[width=1\linewidth, trim=0 0 0 0,clip]{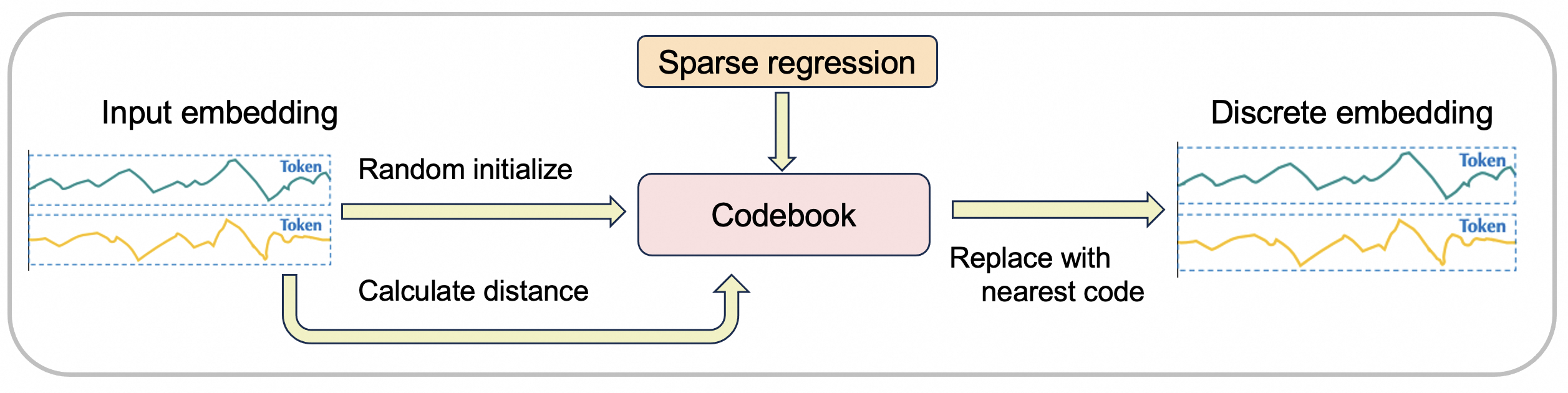}}
\caption{VQ with adaptive codebook}
\label{fig:VQ_adaptive}
% \vskip -0.2in
\end{figure*}

%% file: appendix_vq/appendix_relatedworks.tex
\section{Related Work}
\label{sec_appendix_related_works}

%{\color{blue}

\subsection{Time-series Forecasting}

Temporal variation modeling is a prominent issue in time series analysis and has been extensively investigated with numerous methods been proposed to achieve accurate long-term time series forecasting.

Early studies typically employed traditional statistical methods or machine learning techniques, such as ARIMA~\cite{box_arima2}, Holt-Winter~\cite{Holt-Winter18}, DeepAR~\cite{DBLP:journals/corr/FlunkertSG17-deepAR} and Prophet~\cite{Taylor2018ForecastingAS}. With the rise of deep learning, researchers have turned their attention to more robust and complex neural networks for time series modeling, including TCN~\cite{TCN}, LSTM~\cite{hochreiter_long_1997_lstm}. Specifically, N-BEATS~\cite{nbeats} designed an interpretable layer by encouraging the model to learn trend, seasonality explicitly, and residual components, which shows superior performance on the M4 competition dataset; Dlinear~\cite{Dlinear} employs a simple linear layer, while TimesNet~\cite{TIMESNET} extends the analysis of temporal variations into the 2D space enabling to discover the multi-periodicity adaptively and extract the complex temporal variations. 

Among these, the Transformer~\cite{attention_is_all_you_need} and its subsequent adaptations have demonstrated significant success in long sequence modelling tasks, including time series forecasting. Informer~\cite{haoyietal-informer-2021} proposes a ProbSparse self-attention mechanism and distilling operation to address the quadratic complexity of the Transformer, while Reformer~\cite{KitaevKL20-reformer} replaces dot-product attention by using locality-sensitive hashing and improves its complexity, resulting in significant performance improvements. Subsequently, Autoformer~\cite{Autoformer} designs an efficient auto-correlation mechanism to discover and aggregate information at the series level; FiLM~\cite{zhou2022film} design a Frequency improved Legendre Memory model applying Legendre polynomial projections to approximate historical information, while FEDformer~\cite{FedFormer} proposes an attention mechanism with low-rank approximation in frequency and a mixture of experts decomposition to control the distribution shifting. Additionally, Pyraformer~\cite{liu2022pyraformer} designs pyramidal attention to effectively describe short and long temporal dependencies with low time and space complexity and patchTST~\cite{patchTST} embeds the whole time series of each variate independently into tokens to enlarge local receptive field. GCformer~\cite{GCformer} combines a structured global convolutional branch with a local Transformer-based branch to capture the long and short signals at the same time. Recently, FPT~\cite{zhou2023onefitsall} leverages pretrained language or CV models for time series analysis firstly and achieves excellent performance in all main time series analysis tasks. Overall, the Transformer architecture is widely regarded as one of the most effective and promising approaches for MTS forecasting~\cite{Wen2022transformers,Shao2023exploring}).

\subsection{Distribution shift} 

Although various models above make breakthroughs in time-series forecasting, they often encounter challenges when dealing with non-stationary time-series data, where the distribution of the data changes over time. To address this issue, domain adaptation~\cite{Tzeng17Domain_Adaptation,Ganin16domain_adaptation,Wang18Domain_Adaptation} and domain generalization~\cite{Wang2021GeneralizingTU,Li18DomainGeneralization,pmlr-v28-muandet13,Lu2022DomaininvariantFE}) approaches are commonly employed to mitigate the distribution shift. Domain adaptation algorithms aim to reduce the distribution gap between the source and target domains, while domain generalization algorithms solely rely on the source domain and aim to generalize to the target domain. 
However, defining a domain becomes challenging in the context of non-stationary time series, as the data distribution shifts over time. Recently, \cite{Du2021AdaRNNAL}  proposes the use of Adaptive RNNs to address the distribution shift issues in non-stationary time-series data,  characterizing the distribution information by dividing the training data into periods and then matches the distributions of these identified periods to generalize the model. RevIN~\cite{reversible} utilizes a generally applicable normalization-and-denormalization method with learnable affine transformation to address the distribution shift problem. DIVERSIFY~\cite{lu2023outofdistribution} trys to learn the out of-distribution (OOD) representation on dynamic distributions of times series and then bridges the gap between these latent distributions.

\subsection{Vector Quantization}

Vector quantization (VQ) is a widely used compression technique in signal and image processing, which aims to learn a discrete latent representation by clustering multidimensional data into a finite set of representations. VQ-VAE~\cite{VQVAE} proposes to combine the VQ strategy with a variational autoencoder. There are two key differences between this approach and VAEs: first, the encoder network produces discrete codes instead of continuous ones to obtain a compressed discrete latent space; second, the prior is learned rather than being static; which make it capable of modelling very long term dependencies. SoundStream~\cite{residualVQ} proposes the residual VQ and employs multiple vector quantizers to iteratively quantize the residuals of the waveform.VQ has also been combined with adversarial learning to synthesize high-resolution images, for example, VQGAN ~\cite{VQGAN}. Subsequently, ViT-VQGAN~\cite{VQGAN_use_cosine_sim} proposes to reduce the dimensionality of the codebook and l2-normalize the codebook and TE-VQGAN~\cite{orthogonal} incorporate a regularization term into the loss function to enforce orthogonality among the codebook embeddings. Recent works~\cite{lee2023vq, LENDASSE2005vq, Rasul2022VQARVQ} have applyed Vector Quantization (VQ) to time series domain. Considering veiwing time series as several discrete tokens may be potentially useful, our framework uniquely leverages VQ to enhance the capabilities of transformers in time series forecasting.

%% file: appendix_vq/appendix_supplemental_experiments.tex
\section{Supplemental Experiments}
\label{sec_supplemental_experiments}

\subsection{Dataset Details}
We extensively evaluate the performance of the proposed Sparse-VQ on eight widely used real-world benchmarks, the details of the datasets used in this article are as follows: 1) ETT dataset~\cite{haoyietal-informer-2021} is collected from two separate counties in two versions of the sampling resolution (15 minutes \& 1 h). The ETT dataset contains several time series of electric loads and time series of oil temperature. 2) A dataset called Electricity\footnote{https://archive.ics.uci.edu/ml/datasets/ElectricityLoadDiagrams20112014} contains data on the electricity consumption of more than 300 customers and each column corresponds to the same client. 3) Traffic \footnote{http://pems.dot.ca.gov} dataset records the occupation rate of highway systems in California, USA. 4) The Weather\footnote{https://www.bgc-jena.mpg.de/wetter/} dataset contains 21 meteorological indicators in Germany for an entire year.  5) NorPool \footnote{https://www.nordpoolgroup.com/Market-data1/Power-system-data} includes eight years of hourly energy production volume series in multiple European countries. 6)Wind \cite{li2022generative} contains wind power records from 2020-2021 at 15-minute intervals. 

\begin{figure*}[t]
\centering
\setlength{\abovecaptionskip}{0.cm}
\scalebox{0.80}{
\includegraphics[width=1\linewidth, trim=125 0 125 0,clip]{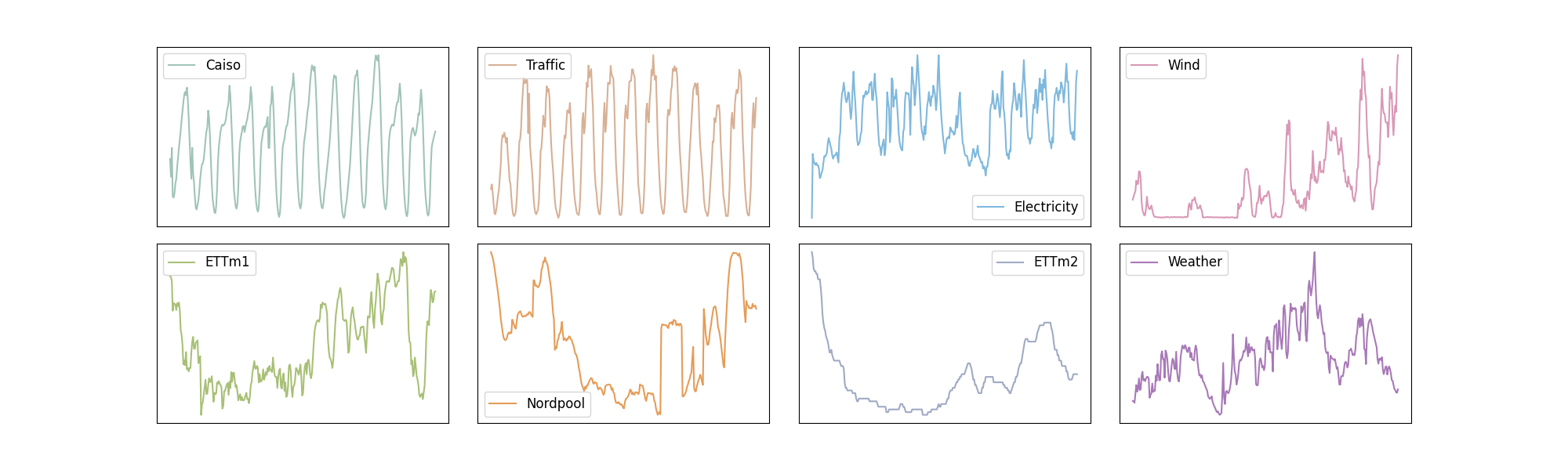}}
\caption{Visualization of the time series of eight datasets for univariate prediction.}
\label{fig:dataset}
%\vskip -0.1in
\end{figure*}

Moreover, we have additionally \href{http://www.energyonline.com/Data}{\textcolor{blue}{gathered}} and processed a novel dataset named \href{https://anonymous.4open.science/r/Sparse-VQ-DC28/dataset/caiso}{\textcolor{blue}{CAISO}} %\footnote{http://www.energyonline.com/Data}
, which contains eight years(2016-2023) of hourly actual electricity load series in different zones of California. Table~\ref{tab:dataset} summarizes all the features of the eight benchmark datasets. We also visualize the time series for univariate prediction in the Figure \ref{fig:dataset} to show different property of these datasets. 

During the experiment, they are divided into training sets, validation sets, and test sets in a 6:2:2 ratio during modeling for ETT and Wind, and 7:1:2 for Weather, Traffic, Electricity, Nordpool and Caiso.

\input{tables_vq/dataset}

\subsection{Implementation Details} 
We use ADAM \cite{kingma_adam:_2017} optimizer with a learning rate of $1e^{-4}$ to $1e^{-3}$. We save models with the lowest loss in validation sets for the final test. Measurements are made using mean square error (MSE) and mean absolute error (MAE). All experiments are repeated 3 times and the mean of the metrics is reported as the final result. Multivariate forecasting results are runed on NVIDIA A100 80GB GPU and other results are runed on NVIDIA V100 32GB GPU. 

\subsection{Long-term Time-series forecasting} 
\label{subsec_appendix_long_forecasting}
Here we verify the consistent performance of our architectural framework on the full ten datasets. To ensure the fairness of the experiments, we follow the classical experiment settings of PatchTST~\cite{patchTST}. Table \ref{tab:appendix_full_bench_uni} shows the full univariate long-term series forecasting results on ten datasets and Table \ref{tab:appendix_ETT_multi} shows multivariate long-term series forecasting results on four ETT datasets. Table \ref{tab:appendix_full_short_forecasting} shows the full results of short-term forecasting.

\subsection{Short-term Time-series forecasting} 
\label{subsec_appendix_short_forecasting}
We conduct short-term forecasting (with relatively short forecasting horizon) experiments on the M4 dataset~\cite{Makridakis2018TheMC}. Table \ref{tab:appendix_full_short_forecasting} shows the full results on marketing data of various frequencies, which show that the performance of Sparse-VQ is superior to OFA and PatchTST, comparable to N-BEATS.

\subsection{Few-shot Time-series forecasting} 
\label{subsec_appendix_few_shot}

During the few-shot forecasting experiment, we only used a certain percentage (5\% in Table \ref{tab:appendix_few_shot_5}, 10\% in Table \ref{tab:appendix_few_shot_10}) timesteps of training data and the evaluation metrics employed are consistent with those used in conventional multivariate time series forecasting. %This experiment was conducted three times, and the average results of these metrics are presented in the subsequent experimental analyses.

\input{appendix_vq/appendix_full_bench_uni}
\input{appendix_vq/appendix_few_shot_5}
\input{appendix_vq/appendix_few_shot_10}
\input{appendix_vq/appendix_ETT_multi}
\input{appendix_vq/appendix_full_short_forecasting}

\subsection{Ablation Experiments} 
\label{subsec_appendix_ablation}

Table~\ref{tab:appendix_modules} shows the full results with and without Sparse-VQ / FFN-free structure. Table~\ref{tab:ablation_vq_boosting} shows the full results of Sparse-VQ cooperated in FEDformer and Autoformer as a plug-in structure. 

\input{tables_vq/ablation_modules}

\input{tables_vq/ablation_vq_boosting}

\subsection{Various Structures of Vector Quantization
}
\label{appendix_various_vq}
We also conducted extensive experiments to explore the the impact of various structures of VQ in Table~\ref{tab:different_vq_structure}.

\input{tables_vq/different_vq_structure}

\subsection{Robustness Analysis} 
\label{subsec_appendix_robustness}

We introduced noise into the data to state-of-the-art models to test their robustness based on the approach described in the MICN [39]. Table~\ref{tab:ablation_noise_VQ} , Table~\ref{tab:appendix_noise_PatchTST} and Table~\ref{tab:appendix_noise_FEDformer} shows the results of Sparse-VQ, PatchTST\cite{patchTST} and FEDformer\cite{FedFormer}. 

\input{tables_vq/ablation_noise_VQ}
\input{appendix_vq/appendix_noise_PatchTST}
\input{appendix_vq/appendix_noise_FEDformer}

%% file: tables_vq/dataset.tex
\begin{table}[h]
%\color{blue}
\caption{Details of ten benchmark datasets.}
\label{tab:dataset}
%\vspace{-2mm}
\vskip -0.2in
\begin{center}
\begin{small}
\begin{sc}
\begin{tabular}{l|cccr}
\toprule
% Dataset & num & dim & freq \\
Dataset & Length & Dimension & Frequency \\
\midrule
ETTm1/m2 & 69680$\approx$2 years & 7 & 15 min\\
ETTh1/h2 & 17420$\approx$2 years & 7 & 1h\\
Electricity & 26304$\approx$3 years & 321 & 1h & \\
Traffic & 17544$\approx$2 years & 862 & 1h & \\
Weather & 52696$\approx$1 years & 21 & 10 min & \\
Wind & 48673$\approx$1 years & 7 & 15min & \\
Nordpool & 70128$\approx$8 years & 18 & 1h & \\
Caiso & 74472$\approx$8 years & 10 & 1h & \\
\midrule

\bottomrule
\end{tabular}
\end{sc}
\end{small}
\end{center}
%\vskip -0.25in
%\vskip -0.1in
\end{table}

%% file: appendix_vq/appendix_full_bench_uni.tex
%\resizebox{!}{\.5\paperheight}{
% \vskip -0.2in
\begin{table*}[h]
\centering
\begin{sc}

% \begin{adjustwidth}{-1.5in}{-1in}
\caption{Univariate long-term series forecasting results on ten datasets with same input length $=512$ and prediction length $\in \{96,192,336,720\}$. A lower MAE indicates better performance. All experiments are repeated 3 times.}\vspace{-1mm}
\scalebox{0.75}{
\begin{tabular}{c|c|cccccccccccccccccc}

\toprule
\multicolumn{2}{c|}{Methods} &\multicolumn{2}{c|}{SVQ} &\multicolumn{2}{c|}{PatchTST} &\multicolumn{2}{c|}{OFA}  &\multicolumn{2}{c|}{Dlinear}  &\multicolumn{2}{c|}{FEDformer} &\multicolumn{2}{c|}{Autoformer} &\multicolumn{2}{c|}{Informer} &\multicolumn{2}{c|}{LogTrans} &\multicolumn{2}{c}{Reformer}\\
\midrule
\multicolumn{2}{c|}{Metric} & MSE  & MAE & MSE & MAE & MSE  & MAE & MSE  & MAE& MSE  & MAE& MSE  & MAE& MSE  & MAE & MSE  & MAE & MSE & MAE\\

\midrule
\multirow{5}{*}{\rotatebox{90}{ETTm1}} 

& 96  & 0.025 & 0.121 & 0.026 & 0.123 & 0.026 & 0.124 & 0.028 & 0.123 & 0.033 & 0.140 & 0.056 & 0.183 & 0.109 & 0.277 & 0.049 & 0.171 & 0.296 & 0.355 \\
 & 192 & 0.039 & 0.150 & 0.040 & 0.151 & 0.040 & 0.153 & 0.045 & 0.156 & 0.058 & 0.186 & 0.081 & 0.216 & 0.151 & 0.31  & 0.157 & 0.317 & 0.429 & 0.474 \\
 & 336 & 0.050 & 0.172 & 0.053 & 0.174 & 0.054 & 0.179 & 0.061 & 0.182 & 0.084 & 0.231 & 0.076 & 0.218 & 0.427 & 0.591 & 0.289 & 0.459 & 0.585 & 0.583 \\
 & 720 & 0.068 & 0.200 & 0.073 & 0.206 & 0.071 & 0.204 & 0.080 & 0.210 & 0.102 & 0.25  & 0.11  & 0.267 & 0.438 & 0.586 & 0.43  & 0.579 & 0.782 & 0.73  \\
 & Avg & \textbf{0.046} & \textbf{0.161} & 0.048 & 0.164 & 0.048 & 0.165 & 0.054 & 0.168 & 0.069 & 0.202 & 0.081 & 0.221 & 0.281 & 0.441 & 0.231 & 0.382 & 0.523 & 0.536 \\

\midrule
\multirow{4}{*}{\rotatebox{90}{ETTm2}}
 & 96  & 0.063 & 0.183 & 0.065 & 0.187 & 0.066 & 0.191 & 0.063 & 0.183 & 0.063 & 0.189 & 0.065 & 0.189 & 0.088 & 0.225 & 0.075 & 0.208 & 0.076 & 0.214 \\
 & 192 & 0.090 & 0.225 & 0.093 & 0.230 & 0.262 & 0.098 & 0.092 & 0.227 & 0.102 & 0.245 & 0.118 & 0.256 & 0.132 & 0.283 & 0.129 & 0.275 & 0.132 & 0.29  \\
 & 336 & 0.118 & 0.262 & 0.122 & 0.267 & 0.305 & 0.134 & 0.119 & 0.261 & 0.13  & 0.279 & 0.154 & 0.305 & 0.18  & 0.336 & 0.154 & 0.302 & 0.16  & 0.312 \\
 & 720 & 0.172 & 0.322 & 0.173 & 0.324 & 0.389 & 0.176 & 0.175 & 0.320 & 0.178 & 0.325 & 0.182 & 0.335 & 0.3   & 0.435 & 0.16  & 0.321 & 0.168 & 0.335 \\
 & Avg & \textbf{ 0.111} & \textbf{0.248} & 0.113 & 0.252 & 0.256 & 0.150 & 0.112 & 0.248 & 0.118 & 0.260 & 0.130 & 0.271 & 0.175 & 0.320 & 0.130 & 0.277 & 0.134 & 0.288 \\

\midrule
\multirow{4}{*}{\rotatebox{90}{ETTh1}} 
 & 96  & 0.056 & 0.184 & 0.059 & 0.189 & 0.061 & 0.192 & 0.056 & 0.180 & 0.079 & 0.215 & 0.071 & 0.206 & 0.193 & 0.377 & 0.283 & 0.468 & 0.532 & 0.569 \\
 & 192 & 0.072 & 0.210 & 0.074 & 0.215 & 0.077 & 0.219 & 0.071 & 0.204 & 0.104 & 0.245 & 0.114 & 0.262 & 0.217 & 0.395 & 0.234 & 0.409 & 0.568 & 0.575 \\
 & 336 & 0.079 & 0.224 & 0.076 & 0.220 & 0.075 & 0.218 & 0.098 & 0.244 & 0.119 & 0.27  & 0.107 & 0.258 & 0.202 & 0.381 & 0.386 & 0.546 & 0.635 & 0.589 \\
 & 720 & 0.084 & 0.231 & 0.087 & 0.236 & 0.090 & 0.240 & 0.189 & 0.359 & 0.142 & 0.299 & 0.126 & 0.283 & 0.183 & 0.355 & 0.475 & 0.628 & 0.762 & 0.666 \\
 & Avg & \textbf{0.073} & \textbf{0.212} & 0.074 & 0.215 & 0.076 & 0.217 & 0.104 & 0.247 & 0.111 & 0.257 & 0.105 & 0.252 & 0.199 & 0.377 & 0.345 & 0.513 & 0.624 & 0.600 \\

\midrule
\multirow{4}{*}{\rotatebox{90}{ETTh2}} 
 & 96  & 0.133 & 0.283 & 0.131 & 0.284 & 0.132 & 0.284 & 0.131 & 0.279 & 0.128 & 0.271 & 0.153 & 0.306 & 0.213 & 0.373 & 0.217 & 0.379 & 1.411 & 0.838 \\
 & 192 & 0.174 & 0.331 & 0.171 & 0.329 & 0.180 & 0.334 & 0.176 & 0.329 & 0.185 & 0.330 & 0.204 & 0.351 & 0.227 & 0.387 & 0.281 & 0.429 & 5.658 & 1.671 \\
 & 336 & 0.181 & 0.343 & 0.171 & 0.336 & 0.179 & 0.338 & 0.209 & 0.367 & 0.231 & 0.378 & 0.246 & 0.389 & 0.242 & 0.401 & 0.293 & 0.437 & 4.777 & 1.582 \\
 & 720 & 0.215 & 0.372 & 0.223 & 0.380 & 0.226 & 0.382 & 0.276 & 0.426 & 0.278 & 0.42  & 0.268 & 0.409 & 0.291 & 0.439 & 0.218 & 0.387 & 2.042 & 1.039 \\
 & Avg & \textbf{0.176} & \textbf{0.332} & 0.174 & 0.332 & 0.179 & 0.335 & 0.198 & 0.350 & 0.206 & 0.350 & 0.218 & 0.364 & 0.243 & 0.4   & 0.252 & 0.408 & 3.472 & 1.283 \\

\midrule
\multirow{5}{*}{\rotatebox{90}{ETTm2}} 

&96  & 0.063           & 0.183           & 0.065  & 0.187  & 0.066  & 0.191  & 0.063  & 0.183          & 0.063  & 0.189  & 0.065  & 0.189 & 0.088  & 0.225 & 0.075  & 0.208 & 0.076  & 0.214 \\
&192 & 0.090           & 0.225           & 0.093  & 0.230  & 0.098  & 0.262  & 0.092  & 0.227          & 0.102  & 0.245  & 0.118  & 0.256 & 0.132  & 0.283 & 0.129  & 0.275 & 0.132  & 0.29  \\
&336 & 0.118           & 0.262           & 0.122  & 0.267  & 0.134  & 0.305  & 0.119  & 0.261          & 0.13   & 0.279  & 0.154  & 0.305 & 0.18   & 0.336 & 0.154  & 0.302 & 0.16   & 0.312 \\
&720 & 0.172           & 0.320           & 0.173  & 0.324  & 0.176  & 0.328  & 0.175  & 0.320          & 0.178  & 0.325  & 0.182  & 0.335 & 0.3    & 0.435 & 0.16   & 0.321 & 0.168  & 0.335 \\
&Avg & \textbf{0.111}  & \textbf{0.248}  & 0.113  & 0.252  & 0.119  & 0.272  & 0.112  & \textbf{0.248} & 0.118  & 0.2595 & 0.130  & 0.271 & 0.175  & 0.320 & 0.130  & 0.277 & 0.134  & 0.288 \\

\midrule
\multirow{5}{*}{\rotatebox{90}{Electricity}}
&96  & 0.194           & 0.304           & 0.209  & 0.321  & 0.212  & 0.321  & 0.202  & 0.314          & 0.253  & 0.37   & 0.341  & 0.438 & 0.484  & 0.538 & 0.288  & 0.393 & 0.274  & 0.379 \\
&192 & 0.226           & 0.325           & 0.246  & 0.344  & 0.266  & 0.358  & 0.233  & 0.336          & 0.282  & 0.386  & 0.345  & 0.428 & 0.557  & 0.558 & 0.432  & 0.483 & 0.304  & 0.402 \\
&336 & 0.263           & 0.358           & 0.280  & 0.379  & 0.306  & 0.385  & 0.270  & 0.364          & 0.346  & 0.431  & 0.406  & 0.47  & 0.636  & 0.613 & 0.43   & 0.483 & 0.37   & 0.448 \\
&720 & 0.298           & 0.404           & 0.332  & 0.424  & 0.383  & 0.455  & 0.303  & 0.406          & 0.422  & 0.484  & 0.565  & 0.581 & 0.819  & 0.682 & 0.491  & 0.531 & 0.46   & 0.511 \\
&Avg & \textbf{0.245}  & \textbf{0.348}  & 0.267  & 0.367  & 0.292  & 0.380  & 0.252  & 0.355          & 0.326  & 0.418  & 0.414  & 0.479 & 0.624  & 0.598 & 0.410  & 0.473 & 0.352  & 0.435 \\

\midrule
\multirow{5}{*}{\rotatebox{90}{Traffic}} 
&96  & 0.115           & 0.183           & 0.134  & 0.223  & 0.145  & 0.252  & 0.122  & 0.194          & 0.17   & 0.263  & 0.246  & 0.346 & 0.257  & 0.353 & 0.226  & 0.317 & 0.313  & 0.383 \\
&192 & 0.112           & 0.182           & 0.130  & 0.221  & 0.148  & 0.255  & 0.125  & 0.199          & 0.173  & 0.265  & 0.266  & 0.37  & 0.299  & 0.376 & 0.314  & 0.408 & 0.386  & 0.453 \\
&336 & 0.113           & 0.190           & 0.133  & 0.227  & 0.156  & 0.261  & 0.124  & 0.202          & 0.178  & 0.266  & 0.263  & 0.371 & 0.312  & 0.387 & 0.387  & 0.453 & 0.423  & 0.468 \\
&720 & 0.126           & 0.204           & 0.146  & 0.241  & 0.173  & 0.276  & 0.139  & 0.222          & 0.187  & 0.286  & 0.269  & 0.372 & 0.366  & 0.436 & 0.491  & 0.437 & 0.378  & 0.433 \\
&Avg & \textbf{0.117}  & \textbf{0.190}  & 0.136  & 0.228  & 0.156  & 0.261  & 0.128  & 0.204          & 0.177  & 0.27   & 0.261  & 0.365 & 0.309  & 0.388 & 0.355  & 0.404 & 0.375  & 0.434 \\

\midrule
\multirow{5}{*}{\rotatebox{90}{Weather}} 
&96  & 0.0009          & 0.0210          & 0.0013 & 0.0265 & 0.0013 & 0.028  & 0.0050 & 0.056          & 0.0035 & 0.046  & 0.011  & 0.081 & 0.0038 & 0.044 & 0.0046 & 0.052 & 0.012  & 0.087 \\
&192 & 0.0011          & 0.0239          & 0.0014 & 0.0281 & 0.0014 & 0.0295 & 0.0061 & 0.065          & 0.0054 & 0.059  & 0.0075 & 0.067 & 0.0023 & 0.04  & 0.0056 & 0.06  & 0.0098 & 0.079 \\
&336 & 0.0013          & 0.0258          & 0.0015 & 0.0289 & 0.0017 & 0.031  & 0.0064 & 0.067          & 0.0041 & 0.05   & 0.0063 & 0.062 & 0.0041 & 0.049 & 0.006  & 0.054 & 0.005  & 0.059 \\
&720 & 0.0018          & 0.0305          & 0.0021 & 0.0341 & 0.0022 & 0.036  & 0.0068 & 0.070          & 0.015  & 0.091  & 0.0085 & 0.07  & 0.0031 & 0.042 & 0.0071 & 0.063 & 0.0041 & 0.049 \\
&Avg & \textbf{0.0013} & \textbf{0.0253} & 0.0016 & 0.0294 & 0.0017 & 0.0311 & 0.0061 & 0.0645         & 0.007  & 0.0615 & 0.0083 & 0.07  & 0.0033 & 0.044 & 0.0058 & 0.057 & 0.0077 & 0.069 \\

\midrule
\multirow{5}{*}{\rotatebox{90}{Wind}} 
&96  & 2.370           & 1.211           & 2.563  & 1.261  & 2.683  & 1.296  & 2.539  & 1.239          & 3.278  & 1.490  & 3.177  & 1.485 & 3.468  & 1.395 & 3.936  & 1.468 & 3.110  & 1.332 \\
&192 & 2.742           & 1.338           & 3.001  & 1.391  & 3.310  & 1.492  & 3.102  & 1.395          & 3.763  & 1.620  & 3.645  & 1.599 & 3.973  & 1.502 & 4.748  & 1.639 & 3.912  & 1.496 \\
&336 & 2.850           & 1.385           & 3.317  & 1.489  & 3.764  & 1.630  & 3.468  & 1.496          & 4.053  & 1.702  & 4.192  & 1.737 & 4.221  & 1.568 & 5.281  & 1.743 & 4.024  & 1.536 \\
&720 & 3.213           & 1.471           & 3.454  & 1.515  & 4.067  & 1.701  & 3.718  & 1.568          & 4.139  & 1.741  & 4.270  & 1.771 & 4.176  & 1.558 & 4.652  & 1.639 & 4.359  & 1.593 \\
&Avg & \textbf{2.794}  & \textbf{1.351}  & 3.084  & 1.414  & 3.456  & 1.530  & 3.207  & 1.425          & 3.808  & 1.638  & 3.821  & 1.648 & 3.960  & 1.506 & 4.654  & 1.622 & 3.851  & 1.489 \\

\midrule
\multirow{5}{*}{\rotatebox{90}{Nordpool}}
&96  & 0.871           & 0.702           & 0.856  & 0.714  & 0.868  & 0.716  & 0.853  & 0.715          & 0.849  & 0.736  & 0.932  & 0.767 & 0.921  & 0.758 & 0.895  & 0.720 & 0.904  & 0.722 \\
&192 & 0.776           & 0.667           & 0.930  & 0.748  & 0.938  & 0.757  & 0.926  & 0.758          & 0.914  & 0.777  & 1.061  & 0.831 & 0.848  & 0.731 & 0.868  & 0.727 & 0.858  & 0.713 \\
&336 & 0.777           & 0.681           & 0.900  & 0.749  & 0.906  & 0.753  & 0.896  & 0.752          & 0.863  & 0.747  & 0.966  & 0.785 & 0.812  & 0.709 & 0.855  & 0.725 & 0.877  & 0.754 \\
&720 & 0.768           & 0.687           & 0.872  & 0.747  & 0.871  & 0.746  & 0.873  & 0.748          & 0.878  & 0.755  & 1.035  & 0.817 & 0.816  & 0.713 & 0.909  & 0.745 & 0.851  & 0.767 \\
&Avg & \textbf{0.798}  & \textbf{0.684}  & 0.890  & 0.740  & 0.896  & 0.743  & 0.887  & 0.743          & 0.876  & 0.754  & 0.999  & 0.8   & 0.849  & 0.728 & 0.882  & 0.729 & 0.873  & 0.739 \\

\midrule
\multirow{5}{*}{\rotatebox{90}{Caiso}}
&96  & 0.140           & 0.255           & 0.162  & 0.279  & 0.169  & 0.285  & 0.151  & 0.262          & 0.191  & 0.327  & 0.227  & 0.353 & 0.217  & 0.337 & 0.206  & 0.311 & 0.207  & 0.340 \\
&192 & 0.211           & 0.308           & 0.212  & 0.322  & 0.233  & 0.328  & 0.215  & 0.312          & 0.247  & 0.367  & 0.262  & 0.370 & 0.293  & 0.394 & 0.254  & 0.360 & 0.221  & 0.344 \\
&336 & 0.257           & 0.342           & 0.263  & 0.355  & 0.281  & 0.361  & 0.259  & 0.346          & 0.266  & 0.360  & 0.299  & 0.399 & 0.302  & 0.405 & 0.310  & 0.422 & 0.298  & 0.396 \\
&720 & 0.325           & 0.391           & 0.325  & 0.403  & 0.373  & 0.424  & 0.346  & 0.404          & 0.372  & 0.451  & 0.504  & 0.535 & 0.382  & 0.464 & 0.380  & 0.479 & 0.356  & 0.441 \\
&Avg & \textbf{0.233}  & \textbf{0.324}  & 0.241  & 0.340  & 0.264  & 0.3495 & 0.243  & 0.331          & 0.269  & 0.376  & 0.323  & 0.414 & 0.299  & 0.4   & 0.288  & 0.393 & 0.271  & 0.380\\

\midrule
% \multicolumn{2}{c|}{Average} & & & & & & & & & & & & & & & \\
\bottomrule
\end{tabular}
\label{tab:appendix_full_bench_uni}
}
% \end{adjustwidth}
\vskip 0.1in
%\vskip -0.1in
\end{sc}
\end{table*}
%}

%% file: appendix_vq/appendix_few_shot_5.tex
%\resizebox{!}{\.5\paperheight}{
% \vskip -0.2in
\begin{table*}[t]
\centering
\begin{sc}

% \begin{adjustwidth}{-1.5in}{-1in}
\caption{Few-shot learning results of four ETT datasets on 5\% data. We use prediction length $O \in \{96, 192, 336, 720\}$.A lower MSE indicates better performance, and the best results are highlighted in bold. ’-’ means that 5\% time series is not sufficient to constitute a training set.}\vspace{-1mm}
\scalebox{0.75}{
\begin{tabular}{c|c|cccccccccccccccccccc}

\toprule
\multicolumn{2}{c|}{Methods}&\multicolumn{2}{c|}{SVQ}&\multicolumn{2}{c|}{OFA}&\multicolumn{2}{c|}{PatchTST}&\multicolumn{2}{c|}{FEDformer}&\multicolumn{2}{c|}{Autoformer}&\multicolumn{2}{c|}{ETSformer}&\multicolumn{2}{c|}{LighTS}&\multicolumn{2}{c|}{Informer}&\multicolumn{2}{c}{Reformer}\\
\midrule
\multicolumn{2}{c|}{Metric} & MSE  & MAE & MSE & MAE & MSE  & MAE & MSE  & MAE& MSE  & MAE& MSE  & MAE& MSE  & MAE & MSE  & MAE & MSE  & MAE \\
\midrule
\multirow{5}{*}{\rotatebox{90}{ETTm1}} 

 & 96  & 0.332          & 0.369          & 0.399 & 0.414 & 0.386 & 0.405 & 0.332 & 0.374 & 0.628 & 0.544 & 0.726 & 0.578 & 1.13  & 0.775 & 1.446 & 0.928 & 1.234 & 0.798 \\
 & 192 & 0.376          & 0.398          & 0.441 & 0.436 & 0.44  & 0.438 & 0.358 & 0.39  & 0.666 & 0.566 & 0.75  & 0.591 & 1.15  & 0.788 & 1.519 & 0.962 & 1.287 & 0.839 \\
 & 336 & 0.425          & 0.423          & 0.499 & 0.467 & 0.485 & 0.459 & 0.402 & 0.416 & 0.807 & 0.628 & 0.851 & 0.659 & 1.198 & 0.809 & 1.774 & 1.032 & 1.288 & 0.842 \\
 & 720 & 0.486          & 0.458          & 0.767 & 0.587 & 0.577 & 0.499 & 0.511 & 0.489 & 0.822 & 0.633 & 0.857 & 0.655 & 1.175 & 0.794 & 1.647 & 0.994 & 1.247 & 0.828 \\
 & Avg & \textbf{0.405} & \textbf{0.412} & 0.526 & 0.476 & 0.472 & 0.45  & 0.400 & 0.417 & 0.73  & 0.592 & 0.796 & 0.62  & 1.163 & 0.791 & 1.597 & 0.979 & 1.264 & 0.826 \\

\midrule
\multirow{4}{*}{\rotatebox{90}{ETTm2}}
 & 96  & 0.190          & 0.274          & 0.206 & 0.288 & 0.199 & 0.28  & 0.236 & 0.326 & 0.229 & 0.32  & 0.232 & 0.322 & 3.599 & 1.478 & 2.119 & 1.189 & 3.883 & 1.545 \\
 & 192 & 0.246          & 0.310          & 0.264 & 0.324 & 0.256 & 0.316 & 0.306 & 0.373 & 0.394 & 0.361 & 0.291 & 0.357 & 3.578 & 1.475 & 2.245 & 1.200 & 3.553 & 1.484 \\
 & 336 & 0.325          & 0.362          & 0.334 & 0.367 & 0.318 & 0.353 & 0.38  & 0.423 & 0.378 & 0.427 & 0.478 & 0.517 & 3.561 & 1.473 & 2.479 & 1.264 & 3.446 & 1.46  \\
 & 720 & 0.413          & 0.409          & 0.454 & 0.432 & 0.46  & 0.436 & 0.674 & 0.583 & 0.523 & 0.51  & 0.553 & 0.538 & 3.896 & 1.533 & 2.846 & 1.320 & 3.445 & 1.46  \\
 & Avg & \textbf{0.294} & \textbf{0.339} & 0.314 & 0.352 & 0.308 & 0.346 & 0.399 & 0.426 & 0.381 & 0.404 & 0.388 & 0.433 & 3.658 & 1.489 & 2.422 & 1.243 & 3.581 & 1.487 \\

\midrule
\multirow{4}{*}{\rotatebox{90}{ETTh1}}
 & 96  & 0.568          & 0.501          & 0.557 & 0.519 & 0.543 & 0.506 & 0.547 & 0.503 & 0.593 & 0.529 & 0.681 & 0.57  & 1.225 & 0.812 & 1.117 & 0.763 & 1.198 & 0.795 \\
 & 192 & 0.734          & 0.560          & 0.711 & 0.57  & 0.748 & 0.58  & 0.72  & 0.604 & 0.652 & 0.563 & 0.725 & 0.602 & 1.249 & 0.828 & 1.376 & 0.860 & 1.273 & 0.853 \\
 & 336 & 0.790          & 0.603          & 0.816 & 0.619 & 0.754 & 0.595 & 0.984 & 0.727 & 0.731 & 0.594 & 0.761 & 0.624 & 1.202 & 0.811 & 1.706 & 0.957 & 1.254 & 0.857 \\
 & 720 & -              & -              & -     & -     & -     & -     & -     & -     & -     & -     & -     & -     & -     & -     & -     & -     & -     & -     \\
 & Avg & \textbf{0.697} & \textbf{0.555} & 0.694 & 0.569 & 0.681 & 0.560 & 0.75  & 0.611 & 0.658 & 0.562 & 0.722 & 0.598 & 1.225 & 0.817 & 1.400 & 0.86  & 1.241 & 0.835 \\

\midrule
\multirow{4}{*}{\rotatebox{90}{ETTh2}} 
 & 96  & 0.429          & 0.452          & 0.401 & 0.421 & 0.376 & 0.421 & 0.442 & 0.456 & 0.39  & 0.424 & 0.428 & 0.468 & 3.837 & 1.508 & 3.345 & 1.478 & 3.753 & 1.518 \\
 & 192 & 0.436          & 0.457          & 0.452 & 0.455 & 0.418 & 0.441 & 0.617 & 0.542 & 0.457 & 0.465 & 0.496 & 0.504 & 3.975 & 1.933 & 3.526 & 1.475 & 3.516 & 1.473 \\
 & 336 & 0.433          & 0.458          & 0.464 & 0.469 & 0.408 & 0.439 & 1.424 & 0.849 & 0.477 & 0.483 & 0.486 & 0.496 & 3.956 & 1.52  & 4.393 & 1.663 & 3.312 & 1.427 \\
 & 720 & -              & -              & -     & -     & -     & -     & -     & -     & -     & -     & -     & -     & -     & -     & -     & -     & -     & -     \\
 & Avg & \textbf{0.433} & \textbf{0.456} & 0.439 & 0.448 & 0.400 & 0.433 & 0.827 & 0.615 & 0.441 & 0.457 & 0.47  & 0.489 & 3.922 & 1.653 & 3.755 & 1.539 & 3.527 & 1.472 \\

\midrule
\multirow{5}{*}{\rotatebox{90}{Electricity}} 
&96  & 0.147  & 0.246   & 0.145 & 0.244 & 0.143 & 0.241 & 0.15  & 0.251 & 0.235 & 0.322 & 0.297 & 0.367 & 1.265 & 0.919 & 0.816 & 0.687 & 1.414 & 0.855 \\
&192 & 0.164  & 0.262   & 0.163 & 0.26  & 0.159 & 0.255 & 0.163 & 0.263 & 0.247 & 0.341 & 0.308 & 0.375 & 1.298 & 0.939 & 0.778 & 0.666 & 1.24  & 0.919 \\
&336 & 0.189  & 0.285   & 0.183 & 0.281 & 0.179 & 0.274 & 0.175 & 0.278 & 0.267 & 0.356 & 0.354 & 0.411 & 1.302 & 0.942 & 0.889 & 0.717 & 1.253 & 0.921 \\
&720 & 0.242  & 0.334   & 0.233 & 0.323 & 0.233 & 0.323 & 0.219 & 0.311 & 0.318 & 0.394 & 0.426 & 0.466 & 1.259 & 0.919 & 1.251 & 0.912 & 1.249 & 0.921 \\
&Avg & \textbf{0.185} & \textbf{0.281} & 0.181 & 0.277 & 0.178 & 0.273 & 0.176 & 0.275 & 0.266 & 0.353 & 0.346 & 0.404 & 1.281 & 0.929 & 0.934 & 0.746 & 1.289 & 0.904 \\

\midrule
\multirow{5}{*}{\rotatebox{90}{Traffic}} 
&96  & 0.413  & 0.279   & 0.404 & 0.286 & 0.419 & 0.298 & 0.427 & 0.304 & 0.67  & 0.421 & 0.795 & 0.481 & 1.557 & 0.821 & 1.149 & 0.599 & 1.586 & 0.841 \\
&192 & 0.425  & 0.287   & 0.412 & 0.294 & 0.434 & 0.305 & 0.447 & 0.315 & 0.653 & 0.405 & 0.837 & 0.503 & 1.596 & 0.834 & 1.247 & 0.657 & 1.602 & 0.844 \\
&336 & 0.440  & 0.297   & 0.439 & 0.31  & 0.449 & 0.313 & 0.478 & 0.333 & 0.707 & 0.445 & 0.867 & 0.523 & 1.621 & 0.841 & 1.531 & 0.799 & 1.668 & 0.868 \\
&720 & -      & -       & -     & -     & -     & -     & -     & -     & -     & -     & -     & -     & -     & -     & -     & -     & -     & -     \\
&Avg & \textbf{0.426}  & \textbf{0.288}   & 0.418 & 0.296 & 0.434 & 0.305 & 0.45  & 0.317 & 0.676 & 0.423 & 0.833 & 0.502 & 1.591 & 0.832 & 1.309 & 0.685 & 1.618 & 0.851 \\

\midrule
\multirow{5}{*}{\rotatebox{90}{Weather}}
&96  & 0.161  & 0.205   & 0.171 & 0.224 & 0.175 & 0.23  & 0.184 & 0.242 & 0.229 & 0.309 & 0.227 & 0.299 & 0.497 & 0.497 & 0.356 & 0.408 & 0.406 & 0.435 \\
&192 & 0.219  & 0.256   & 0.23  & 0.277 & 0.227 & 0.276 & 0.228 & 0.283 & 0.265 & 0.317 & 0.278 & 0.333 & 0.62  & 0.545 & 0.489 & 0.479 & 0.446 & 0.45  \\
&336 & 0.289  & 0.308   & 0.294 & 0.326 & 0.286 & 0.322 & 0.279 & 0.322 & 0.353 & 0.392 & 0.351 & 0.393 & 0.649 & 0.547 & 0.517 & 0.482 & 0.465 & 0.459 \\
&720 & 0.361  & 0.361   & 0.384 & 0.387 & 0.366 & 0.379 & 0.364 & 0.388 & 0.391 & 0.394 & 0.387 & 0.389 & 0.57  & 0.522 & 0.465 & 0.463 & 0.471 & 0.468 \\
&Avg & \textbf{0.258}  & \textbf{0.283}   & 0.269 & 0.303 & 0.263 & 0.301 & 0.263 & 0.308 & 0.309 & 0.353 & 0.31  & 0.353 & 0.584 & 0.527 & 0.457 & 0.458 & 0.447 & 0.453 \\

\midrule
\multirow{5}{*}{\rotatebox{90}{Wind}}
&96  & 1.131  & 0.735   & 1.109 & 0.739 & 1.139 & 0.754 & 1.058 & 0.720 & 1.503 & 0.922 & 1.609 & 0.948 & 5.978 & 1.939 & 4.573 & 1.661 & 2.290 & 1.244 \\
&192 & 1.150  & 0.757   & 1.417 & 0.873 & 1.434 & 0.873 & 1.328 & 0.846 & 1.682 & 0.982 & 1.824 & 1.014 & 4.340 & 1.665 & 4.390 & 1.665 & 2.575 & 1.312 \\
&336 & 1.398  & 0.869   & 1.595 & 0.947 & 1.614 & 0.954 & 1.484 & 0.917 & 1.833 & 1.040 & 2.069 & 1.094 & 3.662 & 1.540 & 3.620 & 1.539 & 2.715 & 1.350 \\
&720 & 1.606  & 0.957   & 1.755 & 1.008 & 1.767 & 1.012 & 1.712 & 0.994 & 1.944 & 1.075 & 2.069 & 1.096 & 2.751 & 1.343 & 2.806 & 1.368 & 2.797 & 1.372 \\
&Avg & \textbf{1.321}  & \textbf{0.830}   & 1.469 & 0.892 & 1.489 & 0.898 & 1.396 & 0.869 & 1.741 & 1.005 & 1.893 & 1.038 & 4.183 & 1.622 & 3.847 & 1.558 & 2.594 & 1.320 \\

\midrule
\multirow{5}{*}{\rotatebox{90}{Nordpool}}
&96  & 0.554  & 0.545   & 0.668 & 0.610 & 0.687 & 0.619 & 0.659 & 0.612 & 0.936 & 0.751 & 0.952 & 0.763 & 4.267 & 1.690 & 2.675 & 1.335 & 1.929 & 1.121 \\
&192 & 0.613  & 0.582   & 0.729 & 0.647 & 0.766 & 0.661 & 0.728 & 0.650 & 0.923 & 0.752 & 1.052 & 0.809 & 3.568 & 1.535 & 2.195 & 1.202 & 2.015 & 1.149 \\
&336 & 0.719  & 0.647   & 0.721 & 0.649 & 0.766 & 0.665 & 0.728 & 0.659 & 0.959 & 0.775 & 1.069 & 0.815 & 2.552 & 1.287 & 1.976 & 1.133 & 2.012 & 1.149 \\
&720 & 0.728  & 0.658   & 0.724 & 0.660 & 0.789 & 0.680 & 0.725 & 0.662 & 0.999 & 0.787 & 0.904 & 0.750 & 2.246 & 1.206 & 1.836 & 1.083 & 2.004 & 1.150 \\
&Avg & \textbf{0.654}  & \textbf{0.608}   & 0.711 & 0.642 & 0.752 & 0.656 & 0.71  & 0.646 & 0.954 & 0.766 & 0.994 & 0.784 & 3.158 & 1.430 & 2.171 & 1.188 & 1.99  & 1.142 \\

\midrule
\multirow{5}{*}{\rotatebox{90}{Caiso}}
&96  & 0.209  & 0.289   & 0.262 & 0.344 & 0.269 & 0.357 & 0.262 & 0.346 & 0.515 & 0.529 & 0.636 & 0.595 & 1.485 & 0.879 & 1.452 & 0.875 & 1.290 & 0.812 \\
&192 & 0.285  & 0.347   & 0.330 & 0.393 & 0.323 & 0.393 & 0.327 & 0.394 & 0.579 & 0.562 & 0.680 & 0.610 & 1.721 & 0.949 & 1.586 & 0.902 & 1.405 & 0.850 \\
&336 & 0.348  & 0.386   & 0.388 & 0.435 & 0.379 & 0.433 & 0.378 & 0.431 & 0.651 & 0.601 & 0.699 & 0.613 & 1.829 & 0.977 & 1.643 & 0.905 & 1.465 & 0.876 \\
&720 & 0.432  & 0.442   & 0.564 & 0.533 & 0.558 & 0.539 & 0.560 & 0.534 & 0.866 & 0.703 & 0.901 & 0.712 & 2.106 & 1.057 & 1.975 & 1.006 & 1.573 & 0.909 \\
&Avg & \textbf{0.319}  & \textbf{0.366}   & 0.386 & 0.426 & 0.382 & 0.431 & 0.382 & 0.426 & 0.653 & 0.599 & 0.729 & 0.633 & 1.785 & 0.966 & 1.664 & 0.922 & 1.433 & 0.862\\

\midrule

% \midrule
% \multicolumn{2}{c|}{Average} & & & & & & & & & & & & & & & \\
\bottomrule
\end{tabular}
\label{tab:appendix_few_shot_5}
}
% \end{adjustwidth}
\vskip 0.1in
%\vskip -0.1in
\end{sc}
\end{table*}
%}

%% file: appendix_vq/appendix_few_shot_10.tex
%\resizebox{!}{\.5\paperheight}{
% \vskip -0.2in
\begin{table*}[h]
\centering
\begin{sc}

% \begin{adjustwidth}{-1.5in}{-1in}
\caption{Few-shot learning results of four ETT datasets on 10\% data. We use prediction length $O \in \{96, 192, 336, 720\}$.A lower MSE indicates better performance, and the best results are highlighted in bold. ’-’ means that 10\% time series is not sufficient to constitute a training set.}\vspace{-1mm}
\scalebox{0.75}{
\begin{tabular}{c|c|cccccccccccccccccccc}

\toprule
\multicolumn{2}{c|}{Methods}&\multicolumn{2}{c|}{SVQ}&\multicolumn{2}{c|}{OFA}&\multicolumn{2}{c|}{PatchTST}&\multicolumn{2}{c|}{FEDformer}&\multicolumn{2}{c|}{Autoformer}&\multicolumn{2}{c|}{ETSformer}&\multicolumn{2}{c|}{LighTS}&\multicolumn{2}{c|}{Informer}&\multicolumn{2}{c}{Reformer}\\
\midrule
\multicolumn{2}{c|}{Metric} & MSE  & MAE & MSE & MAE & MSE  & MAE & MSE  & MAE& MSE  & MAE& MSE  & MAE& MSE  & MAE & MSE  & MAE & MSE  & MAE \\
\midrule
\multirow{5}{*}{\rotatebox{90}{ETTm1}} 

& 96  & 0.333          & 0.365          & 0.41  & 0.419 & 0.39  & 0.404 & 0.352 & 0.392 & 0.578 & 0.518 & 0.774 & 0.614 & 1.162 & 0.785 & 1.555 & 0.910 & 1.442 & 0.847 \\
 & 192 & 0.370          & 0.388          & 0.437 & 0.434 & 0.429 & 0.423 & 0.382 & 0.412 & 0.617 & 0.546 & 0.754 & 0.592 & 1.172 & 0.793 & 1.883 & 1.033 & 1.444 & 0.862 \\
 & 336 & 0.413          & 0.416          & 0.476 & 0.454 & 0.469 & 0.439 & 0.419 & 0.434 & 0.998 & 0.775 & 0.869 & 0.677 & 1.227 & 0.908 & 2.095 & 1.110 & 1.45  & 0.866 \\
 & 720 & 0.476          & 0.452          & 0.681 & 0.556 & 0.569 & 0.498 & 0.49  & 0.477 & 0.693 & 0.579 & 0.81  & 0.63  & 1.207 & 0.797 & 2.389 & 1.167 & 1.366 & 0.85  \\
 & Avg & \textbf{0.398} & \textbf{0.405} & 0.501 & 0.466 & 0.464 & 0.441 & 0.411 & 0.429 & 0.722 & 0.605 & 0.802 & 0.628 & 1.192 & 0.821 & 1.981 & 1.055 & 1.426 & 0.856 \\
 
\midrule
\multirow{4}{*}{\rotatebox{90}{ETTm2}} 
 & 96  & 0.174          & 0.256          & 0.191 & 0.274 & 0.188 & 0.269 & 0.213 & 0.303 & 0.291 & 0.399 & 0.352 & 0.454 & 3.203 & 1.407 & 2.185 & 1.167 & 4.195 & 1.628 \\
 & 192 & 0.235          & 0.297          & 0.252 & 0.317 & 0.251 & 0.309 & 0.278 & 0.345 & 0.307 & 0.379 & 0.694 & 0.691 & 3.112 & 1.387 & 2.509 & 1.242 & 4.042 & 1.601 \\
 & 336 & 0.292          & 0.336          & 0.306 & 0.353 & 0.307 & 0.346 & 0.338 & 0.385 & 0.543 & 0.559 & 2.408 & 1.407 & 3.255 & 1.421 & 2.336 & 1.223 & 3.963 & 1.585 \\
 & 720 & 0.394          & 0.397          & 0.433 & 0.427 & 0.426 & 0.417 & 0.436 & 0.44  & 0.712 & 0.614 & 1.913 & 1.166 & 3.909 & 1.543 & 3.325 & 1.446 & 3.711 & 1.532 \\
 & Avg & \textbf{0.274} & \textbf{0.322} & 0.296 & 0.343 & 0.29  & 0.335 & 0.316 & 0.368 & 0.463 & 0.488 & 1.342 & 0.930 & 3.370 & 1.440 & 2.589 & 1.270 & 3.978 & 1.587 \\

\midrule
\multirow{4}{*}{\rotatebox{90}{ETTh1}} 
 & 96  & 0.429          & 0.441          & 0.516 & 0.485 & 0.458 & 0.456 & 0.492 & 0.495 & 0.512 & 0.499 & 0.613 & 0.552 & 1.179 & 0.792 & 1.523 & 0.938 & 1.184 & 0.79  \\
 & 192 & 0.471          & 0.468          & 0.598 & 0.524 & 0.57  & 0.516 & 0.565 & 0.538 & 0.624 & 0.555 & 0.722 & 0.598 & 1.199 & 0.806 & 1.572 & 0.929 & 1.295 & 0.85  \\
 & 336 & 0.578          & 0.531          & 0.657 & 0.55  & 0.608 & 0.535 & 0.721 & 0.622 & 0.691 & 0.574 & 0.75  & 0.619 & 1.202 & 0.811 & 1.593 & 0.914 & 1.294 & 0.854 \\
 & 720 & 0.827          & 0.641          & 0.762 & 0.61  & 0.725 & 0.591 & 0.986 & 0.743 & 0.728 & 0.614 & 0.721 & 0.616 & 1.217 & 0.825 & 1.843 & 0.995 & 1.223 & 0.838 \\
 & Avg & \textbf{0.576} & \textbf{0.520} & 0.633 & 0.542 & 0.590 & 0.525 & 0.691 & 0.6   & 0.639 & 0.561 & 0.702 & 0.596 & 1.199 & 0.809 & 1.633 & 0.944 & 1.249 & 0.833 \\

\midrule
\multirow{4}{*}{\rotatebox{90}{ETTh2}}
 & 96  & 0.294          & 0.358          & 0.353 & 0.389 & 0.331 & 0.374 & 0.357 & 0.411 & 0.382 & 0.416 & 0.413 & 0.451 & 3.837 & 1.508 & 3.076 & 1.385 & 3.788 & 1.533 \\
 & 192 & 0.357          & 0.395          & 0.403 & 0.414 & 0.402 & 0.411 & 0.569 & 0.519 & 0.478 & 0.474 & 0.474 & 0.477 & 3.856 & 1.513 & 3.608 & 1.504 & 3.552 & 1.483 \\
 & 336 & 0.386          & 0.422          & 0.426 & 0.441 & 0.406 & 0.433 & 0.671 & 0.572 & 0.504 & 0.501 & 0.547 & 0.543 & 3.952 & 1.526 & 3.542 & 1.497 & 3.395 & 1.526 \\
 & 720 & 0.467          & 0.482          & 0.477 & 0.48  & 0.449 & 0.464 & 0.824 & 0.648 & 0.499 & 0.509 & 0.516 & 0.523 & 3.842 & 1.503 & 4.443 & 1.697 & 3.205 & 1.401 \\
 & Avg & \textbf{0.376} & \textbf{0.414} & 0.415 & 0.431 & 0.397 & 0.421 & 0.605 & 0.538 & 0.466 & 0.475 & 0.488 & 0.499 & 3.872 & 1.513 & 3.667 & 1.521 & 3.485 & 1.486 \\

\midrule
\multirow{5}{*}{\rotatebox{90}{Electricity}} 
 & 96  & 0.145          & 0.243          & 0.14  & 0.238 & 0.139 & 0.237  & 0.15  & 0.253 & 0.231 & 0.323  & 0.261 & 0.348 & 1.259 & 0.919 & 0.649 & 0.584 & 0.993 & 0.784 \\
 & 192 & 0.162          & 0.259          & 0.16  & 0.255 & 0.156 & 0.252  & 0.164 & 0.264 & 0.261 & 0.356  & 0.338 & 0.406 & 1.16  & 0.873 & 0.667 & 0.594 & 0.938 & 0.753 \\
 & 336 & 0.188          & 0.284          & 0.18  & 0.276 & 0.175 & 0.27   & 0.181 & 0.282 & 0.36  & 0.445  & 0.41  & 0.474 & 1.157 & 0.872 & 0.711 & 0.620 & 0.925 & 0.745 \\
 & 720 & 0.245          & 0.325          & 0.241 & 0.323 & 0.233 & 0.317  & 0.223 & 0.321 & 0.53  & 0.585  & 0.715 & 0.685 & 1.203 & 0.898 & 0.833 & 0.690 & 1.004 & 0.79  \\
 & Avg & \textbf{0.185} & \textbf{0.278} & 0.18  & 0.273 & 0.176 & 0.269  & 0.18  & 0.28  & 0.346 & 0.427  & 0.431 & 0.478 & 1.195 & 0.891 & 0.715 & 0.622 & 0.965 & 0.768 \\

\midrule
\multirow{5}{*}{\rotatebox{90}{Traffic}}
 & 96  & 0.409          & 0.272          & 0.403 & 0.289 & 0.414 & 0.297  & 0.419 & 0.298 & 0.639 & 0.4    & 0.672 & 0.405 & 1.557 & 0.821 & 0.928 & 0.513 & 1.527 & 0.815 \\
 & 192 & 0.423          & 0.275          & 0.415 & 0.296 & 0.426 & 0.301  & 0.434 & 0.305 & 0.637 & 0.416  & 0.727 & 0.424 & 1.454 & 0.765 & 0.976 & 0.530 & 1.538 & 0.817 \\
 & 336 & 0.428          & 0.279          & 0.426 & 0.304 & 0.434 & 0.303  & 0.449 & 0.313 & 0.655 & 0.427  & 0.749 & 0.454 & 1.521 & 0.812 & 1.015 & 0.545 & 1.55  & 0.819 \\
 & 720 & 0.445          & 0.296          & 0.474 & 0.331 & 0.487 & 0.337  & 0.484 & 0.336 & 0.722 & 0.456  & 0.847 & 0.499 & 1.605 & 0.846 & 1.162 & 0.607 & 1.588 & 0.833 \\
 & Avg & \textbf{0.426} & \textbf{0.281} & 0.430 & 0.305 & 0.44  & 0.31   & 0.447 & 0.313 & 0.663 & 0.425  & 0.749 & 0.446 & 1.534 & 0.811 & 1.020 & 0.549 & 1.551 & 0.821 \\

\midrule
\multirow{5}{*}{\rotatebox{90}{Weather}} 
& 96  & 0.152          & 0.196          & 0.165 & 0.215 & 0.163 & 0.215  & 0.171 & 0.224 & 0.188 & 0.253  & 0.221 & 0.297 & 0.374 & 0.401 & 0.305 & 0.371 & 0.335 & 0.38  \\
 & 192 & 0.201          & 0.241          & 0.21  & 0.257 & 0.21  & 0.254  & 0.215 & 0.263 & 0.25  & 0.304  & 0.27  & 0.322 & 0.552 & 0.478 & 0.425 & 0.432 & 0.522 & 0.462 \\
 & 336 & 0.252          & 0.282          & 0.259 & 0.297 & 0.256 & 0.292  & 0.258 & 0.299 & 0.312 & 0.346  & 0.32  & 0.351 & 0.724 & 0.541 & 0.605 & 0.500 & 0.715 & 0.535 \\
 & 720 & 0.325          & 0.335          & 0.332 & 0.346 & 0.321 & 0.339  & 0.32  & 0.346 & 0.387 & 0.393  & 0.39  & 0.396 & 0.739 & 0.558 & 0.714 & 0.536 & 0.611 & 0.5   \\
 & Avg & \textbf{0.233} & \textbf{0.264} & 0.242 & 0.279 & 0.238 & 0.275  & 0.241 & 0.283 & 0.284 & 0.324  & 0.3   & 0.342 & 0.597 & 0.495 & 0.512 & 0.460 & 0.546 & 0.469 \\

\midrule
\multirow{5}{*}{\rotatebox{90}{Wind}}  
 & 96  & 0.907          & 0.627          & 1.070 & 0.717 & 1.075 & 0.718  & 1.012 & 0.697 & 1.490 & 0.911  & 1.464 & 0.905 & 5.424 & 1.935 & 3.858 & 1.649 & 2.570 & 1.331 \\
 & 192 & 1.150          & 0.757          & 1.331 & 0.836 & 1.342 & 0.843  & 1.266 & 0.820 & 1.671 & 0.977  & 1.850 & 1.047 & 4.793 & 1.823 & 3.062 & 1.440 & 2.953 & 1.439 \\
 & 336 & 1.398          & 0.869          & 1.574 & 0.937 & 1.545 & 0.928  & 1.474 & 0.911 & 1.822 & 1.041  & 1.976 & 1.101 & 4.313 & 1.740 & 3.111 & 1.467 & 2.923 & 1.428 \\
 & 720 & 1.606          & 0.957          & 1.743 & 1.005 & 1.749 & 1.004  & 1.635 & 0.975 & 1.836 & 1.056  & 1.947 & 1.084 & 3.976 & 1.662 & 3.069 & 1.456 & 2.975 & 1.440 \\
 & Avg & \textbf{1.265} & \textbf{0.803} & 1.430 & 0.874 & 1.428 & 0.873  & 1.347 & 0.851 & 1.705 & 0.996  & 1.809 & 1.034 & 4.627 & 1.790 & 3.275 & 1.503 & 2.855 & 1.410 \\

\midrule
\multirow{5}{*}{\rotatebox{90}{Nordpool}} 
 & 96  & 0.554          & 0.545          & 0.631 & 0.591 & 0.638 & 0.594  & 0.629 & 0.595 & 0.789 & 0.694  & 0.888 & 0.737 & 2.247 & 1.190 & 1.889 & 1.116 & 1.876 & 1.102 \\
 & 192 & 0.613          & 0.582          & 0.707 & 0.636 & 0.704 & 0.633  & 0.687 & 0.635 & 0.734 & 0.673  & 1.031 & 0.798 & 2.400 & 1.240 & 2.052 & 1.141 & 2.055 & 1.140 \\
 & 336 & 0.594          & 0.581          & 0.674 & 0.626 & 0.679 & 0.627  & 0.678 & 0.635 & 0.756 & 0.680  & 0.860 & 0.727 & 2.484 & 1.267 & 2.022 & 1.121 & 2.175 & 1.175 \\
 & 720 & 0.577          & 0.578          & 0.660 & 0.626 & 0.674 & 0.632  & 0.677 & 0.638 & 0.812 & 0.707  & 0.938 & 0.758 & 2.579 & 1.385 & 1.641 & 1.013 & 2.161 & 1.178 \\
 & Avg & \textbf{0.585} & \textbf{0.572} & 0.668 & 0.620 & 0.674 & 0.6215 & 0.668 & 0.626 & 0.773 & 0.689  & 0.929 & 0.755 & 2.428 & 1.271 & 1.901 & 1.098 & 2.067 & 1.149 \\

\midrule
\multirow{5}{*}{\rotatebox{90}{Caiso}} 
 & 96  & 0.210          & 0.290          & 0.246 & 0.328 & 0.243 & 0.329  & 0.247 & 0.329 & 0.499 & 0.514  & 0.664 & 0.607 & 1.641 & 0.902 & 1.229 & 0.808 & 1.226 & 0.790 \\
 & 192 & 0.285          & 0.349          & 0.320 & 0.382 & 0.317 & 0.382  & 0.315 & 0.380 & 0.539 & 0.534  & 0.835 & 0.671 & 1.801 & 0.945 & 1.269 & 0.820 & 1.317 & 0.822 \\
 & 336 & 0.334          & 0.384          & 0.377 & 0.421 & 0.376 & 0.421  & 0.365 & 0.413 & 0.634 & 0.583  & 0.795 & 0.655 & 2.076 & 1.026 & 1.442 & 0.858 & 1.401 & 0.846 \\
 & 720 & 0.436          & 0.444          & 0.525 & 0.509 & 0.547 & 0.523  & 0.494 & 0.499 & 0.807 & 0.667  & 0.884 & 0.705 & 2.361 & 1.230 & 1.560 & 0.891 & 1.735 & 0.942 \\
 & Avg & \textbf{0.316} & \textbf{0.367} & 0.367 & 0.41  & 0.371 & 0.414  & 0.355 & 0.405 & 0.620 & 0.5745 & 0.795 & 0.660 & 1.970 & 1.026 & 1.375 & 0.844 & 1.420 & 0.85\\ 
 
\midrule

% \midrule
% \multicolumn{2}{c|}{Average} & & & & & & & & & & & & & & & \\
\bottomrule
\end{tabular}
\label{tab:appendix_few_shot_10}
}
% \end{adjustwidth}
\vskip 0.1in
%\vskip -0.1in
\end{sc}
\end{table*}
%}

%% file: appendix_vq/appendix_ETT_multi.tex
%\resizebox{!}{\.5\paperheight}{
% \vskip -0.2in
\begin{table*}[h]
\centering
\begin{sc}

% \begin{adjustwidth}{-1.5in}{-1in}
\caption{ Multivariate long-term series forecasting results on four ETT datasets with same input length $=512$ and various prediction length $\in \{96,192,336,720\}$ . A lower MAE indicates better performance. All experiments are repeated 3 times.}\vspace{-1mm}
\scalebox{0.85}{
\begin{tabular}{c|c|cccccccccccccccccc}

\toprule
\multicolumn{2}{c|}{Methods}&\multicolumn{2}{c|}{SVQ}&\multicolumn{2}{c|}{OFA}&\multicolumn{2}{c|}{PatchTST}&\multicolumn{2}{c|}{FEDformer}&\multicolumn{2}{c|}{Autoformer}&\multicolumn{2}{c|}{ETSformer}&\multicolumn{2}{c|}{LighTS}&\multicolumn{2}{c|}{Informer}&\multicolumn{2}{c}{Reformer}\\
\midrule
\multicolumn{2}{c|}{Metric} & MSE  & MAE & MSE & MAE & MSE  & MAE & MSE  & MAE& MSE  & MAE& MSE  & MAE& MSE  & MAE & MSE  & MAE & MSE & MAE\\
\midrule
\multirow{5}{*}{\rotatebox{90}{ETTm1}} 

 & 96  & 0.284          & 0.329          & 0.293 & 0.346 & 0.292 & 0.346 & 0.299 & 0.343 & 0.379 & 0.419 & 0.505 & 0.475 & 0.672 & 0.571 & 0.6   & 0.546 & 0.538 & 0.528 \\
 & 192 & 0.329          & 0.355          & 0.333 & 0.370 & 0.332 & 0.372 & 0.335 & 0.365 & 0.426 & 0.441 & 0.553 & 0.496 & 0.795 & 0.669 & 0.837 & 0.7   & 0.658 & 0.592 \\
 & 336 & 0.364          & 0.376          & 0.369 & 0.392 & 0.366 & 0.394 & 0.369 & 0.386 & 0.445 & 0.459 & 0.621 & 0.537 & 1.212 & 0.871 & 1.124 & 0.832 & 0.898 & 0.721 \\
 & 720 & 0.422          & 0.409          & 0.416 & 0.420 & 0.417 & 0.421 & 0.425 & 0.421 & 0.543 & 0.490 & 0.671 & 0.561 & 1.166 & 0.823 & 1.153 & 0.82  & 1.102 & 0.841 \\
 & Avg & \textbf{0.350} & \textbf{0.367} & 0.353 & 0.382 & 0.352 & 0.383 & 0.357 & 0.379 & 0.448 & 0.452 & 0.588 & 0.517 & 0.961 & 0.734 & 0.929 & 0.725 & 0.799 & 0.671 \\

\midrule
\multirow{4}{*}{\rotatebox{90}{ETTm2}}
 & 96  & 0.159          & 0.243          & 0.166 & 0.256 & 0.173 & 0.262 & 0.167 & 0.26  & 0.203 & 0.287 & 0.255 & 0.339 & 0.705 & 0.69  & 0.768 & 0.642 & 0.365 & 0.453 \\
 & 192 & 0.216          & 0.283          & 0.223 & 0.296 & 0.229 & 0.301 & 0.224 & 0.303 & 0.269 & 0.328 & 0.281 & 0.34  & 0.924 & 0.692 & 0.989 & 0.757 & 0.533 & 0.563 \\
 & 336 & 0.268          & 0.317          & 0.274 & 0.329 & 0.286 & 0.341 & 0.281 & 0.342 & 0.325 & 0.366 & 0.339 & 0.372 & 1.364 & 0.877 & 1.334 & 0.872 & 1.363 & 0.887 \\
 & 720 & 0.349          & 0.371          & 0.362 & 0.385 & 0.378 & 0.401 & 0.397 & 0.421 & 0.421 & 0.415 & 0.422 & 0.419 & 0.877 & 1.074 & 3.048 & 1.328 & 3.379 & 1.338 \\
 & Avg & \textbf{0.248} & \textbf{0.304} & 0.256 & 0.317 & 0.267 & 0.326 & 0.267 & 0.332 & 0.305 & 0.349 & 0.324 & 0.368 & 0.968 & 0.833 & 1.535 & 0.900 & 1.41  & 0.810 \\

\midrule
\multirow{4}{*}{\rotatebox{90}{ETTh1}}
 & 96  & 0.358          & 0.385          & 0.370 & 0.400 & 0.376 & 0.397 & 0.375 & 0.399 & 0.376 & 0.419 & 0.449 & 0.459 & 0.865 & 0.713 & 0.878 & 0.74  & 0.837 & 0.728 \\
 & 192 & 0.401          & 0.416          & 0.413 & 0.429 & 0.416 & 0.418 & 0.405 & 0.416 & 0.420 & 0.448 & 0.5   & 0.482 & 1.008 & 0.792 & 1.037 & 0.824 & 0.923 & 0.766 \\
 & 336 & 0.425          & 0.435          & 0.422 & 0.440 & 0.442 & 0.433 & 0.439 & 0.443 & 0.459 & 0.465 & 0.521 & 0.496 & 1.107 & 0.809 & 1.238 & 0.932 & 1.097 & 0.835 \\
 & 720 & 0.438          & 0.459          & 0.447 & 0.468 & 0.477 & 0.456 & 0.472 & 0.490 & 0.506 & 0.507 & 0.514 & 0.512 & 1.181 & 0.865 & 1.135 & 0.852 & 1.257 & 0.889 \\
 & Avg & \textbf{0.406} & \textbf{0.424} & 0.413 & 0.434 & 0.428 & 0.426 & 0.423 & 0.437 & 0.440 & 0.460 & 0.496 & 0.487 & 1.040 & 0.795 & 1.072 & 0.837 & 1.029 & 0.805 \\

\midrule
\multirow{4}{*}{\rotatebox{90}{ETTh2}} 
 & 96  & 0.272          & 0.330          & 0.274 & 0.337 & 0.285 & 0.342 & 0.289 & 0.353 & 0.346 & 0.388 & 0.358 & 0.397 & 3.755 & 1.525 & 2.116 & 1.197 & 2.626 & 1.317 \\
 & 192 & 0.331          & 0.371          & 0.341 & 0.382 & 0.354 & 0.389 & 0.383 & 0.418 & 0.429 & 0.439 & 0.456 & 0.452 & 5.602 & 1.931 & 4.315 & 1.635 & 11.12 & 2.979 \\
 & 336 & 0.344          & 0.394          & 0.329 & 0.384 & 0.373 & 0.407 & 0.448 & 0.465 & 0.496 & 0.487 & 0.482 & 0.486 & 4.721 & 1.835 & 1.124 & 1.604 & 9.323 & 2.769 \\
 & 720 & 0.387          & 0.426          & 0.379 & 0.422 & 0.406 & 0.441 & 0.605 & 0.551 & 0.463 & 0474  & 0.515 & 0.511 & 3.647 & 1.625 & 3.188 & 1.54  & 3.874 & 1.697 \\
 & Avg & \textbf{0.334} & \textbf{0.380} & 0.331 & 0.381 & 0.355 & 0.395 & 0.431 & 0.447 & 0.434 & 0.438 & 0.453 & 0.462 & 4.431 & 1.729 & 2.686 & 1.494 & 6.736 & 2.191 \\

\midrule
% \midrule
% \multicolumn{2}{c|}{Average} & & & & & & & & & & & & & & & \\
\bottomrule
\end{tabular}
\label{tab:appendix_ETT_multi}
}
% \end{adjustwidth}
\vskip 0.2in
%\vskip -0.1in
\end{sc}
\end{table*}
%}

%% file: appendix_vq/appendix_full_short_forecasting.tex
%\resizebox{!}{\.5\paperheight}{
% \vskip -0.2in
\begin{table*}[t]
\centering
\begin{sc}

% \begin{adjustwidth}{-1.5in}{-1in}
\caption{Short-term forecasting task on M4. The prediction lengths are $\in \{6,48\}$. A lower score indicates better performance. All experiments are repeated 3 times.} \vspace{-1mm}
\scalebox{0.85}{
\begin{tabular}{c|c|cccccccccccccccccc}

\toprule
\multicolumn{2}{c|}{Methods} &{SVQ} &{OFA} &{PatchTST} &{N-HiTS} &{N-BEATS} &{ETSformer} &{LighTS} &{Dlinear} &{FEDformer} &{Autoformer} &{Informer} &{Reformer} \\

\midrule
\multirow{3}{*}{\rotatebox{90}{Yearly}} 

& SMAPE & {13.279} & {13.531} & 13.477 & 13.418 &13.436 &18.009 &14.247 & 16.965 &13.728 & 13.974 & 14.727 & 16.169 \\
& MASE & {2.974} & {3.0154} & 3.019 &3.045 &3.043 &4.487 &3.109 &4.283 &3.048 & 3.134 &3.418 &3.800  \\
& OWA & {0.78} & {0.793} & 0.792 &0.793 &0.794 &1.115 &0.827 &1.058 &0.803 & 0.822 &0.881 &0.973   \\

\midrule
\multirow{3}{*}{\rotatebox{90}{Quarterly}} 

& SMAPE & {10.118} & {10.177} &10.38  &10.202 &10.124 &13.376 &11.364 &12.145 &10.792 &11.338 &11.360 &13.313  \\[1.1ex]
& MASE & {1.181} & {1.194} &1.233  &1.194 &1.169 &1.906 &1.328 &1.520 &1.283 &1.365 &1.401 &1.775   \\[1.1ex]
& OWA & {0.89} & {0.898} & 0.921 &0.899 &0.886 &1.302 &1.000 &1.106 &0.958 &1.012 &1.027 &1.252 \\[1.1ex]
\midrule

\multirow{3}{*}{\rotatebox{90}{Monthly}} 

&SMAPE  & 12.929 & {12.894} & 12.959 &12.791 &12.677 &14.588 &14.014 &13.514 &14.260 &13.958 &14.062 &20.128   \\[1.1ex]
& MASE & 0.964 & {0.956} & 0.97 &0.969 &0.937 &1.368 &1.053 &1.037 &1.102 &1.103 &1.141 &2.614   \\[1.1ex]
& OWA & 0.901 & {0.897} & 0.905 &0.899 &0.880 &1.149 &0.981 &0.956 &1.012 &1.002 &1.024 &1.927 \\[1.1ex]

\midrule

\multirow{3}{*}{\rotatebox{90}{Others}} 

& SMAPE & {4.985} & {4.940} & 4.952 &5.061 &4.925 &7.267 &15.880 &6.709 &4.954 &5.485 &24.460 &32.491  \\
& MASE & {3.248} & {3.228} &3.347 &3.216 &3.391 &5.240 &11.434 &4.953 &3.264 &3.865 &20.960 &33.355   \\
& OWA & {1.037} & {1.029} &1.049 &1.040 &1.053 &1.591 &3.474 &1.487 &1.036 &1.187 &5.879 &8.679  \\
\midrule

\multirow{3}{*}{\rotatebox{90}{Average}} 

& SMAPE & {11.938} &{11.991} &12.059&11.927 &11.851 &14.718 &13.525 &13.639 &12.840 &12.909 &14.086 &18.200 \\
& MASE & {1.593} & {1.600} &1.623 &1.613 &1.599 &2.408 &2.111 &2.095 &1.701 &1.771 &2.718 &4.223   \\
& OWA & {0.857} & {0.861} &0.869 &0.861 &0.855 &1.172 &1.051 &1.051 &0.918&0.939 &1.230 &1.775 \\
\midrule

% \midrule
% \multicolumn{2}{c|}{Average} & & & & & & & & & & & & & & & \\
\bottomrule
\end{tabular}
\label{tab:appendix_full_short_forecasting}
}
% \end{adjustwidth}
\vskip 0.1in
%\vskip -0.1in
\end{sc}
\end{table*}
%}

%% file: tables_vq/ablation_modules.tex
%\resizebox{!}{\.5\paperheight}{
% \vskip -0.2in
\begin{table*}[h]
\centering

%\begin{footnotesize}
% \begin{adjustwidth}{-1.5in}{-1in}
\caption{Ablation study of FFN-free and Sparse-VQ in PatchTST. 4 cases are included: (a) both FFN-free and Sparse-VQ are included in model (SVQ+FFN-f); (b) only Vector Quantization (VQ+FFN-f); (c) only FFN-free(FFN-f);(d) neither of them is included (Original patchTST model). The best results are in bold. A lower MSE indicates better performance. All experiments are repeated 3 times.}\vspace{-3mm}

\begin{center}
% \begin{small}
\begin{sc}
\scalebox{1.0}{
\begin{tabular}{c|c|cccccccccccccccc}
\toprule
% \multicolumn{2}{c|}{Methods}&\multicolumn{2}{c|}{PatchTST(96)}&\multicolumn{2}{c|}{Attention}&\multicolumn{2}{c|}{Series-LG/GL}&\multicolumn{2}{c|}{Series-GL}&\multicolumn{2}{c}{Concatenate}\\
\multicolumn{2}{c|}{Methods}&\multicolumn{2}{c|}{SVQ+FFN-f}&\multicolumn{2}{c|}{VQ+FFN-f}&\multicolumn{2}{c|}{FFN-f}&\multicolumn{2}{c}{Original}\\
\midrule
\multicolumn{2}{c|}{Metric} & MSE  & MAE & MSE & MAE& MSE  & MAE& MSE  & MAE\\
\midrule
\multirow{4}{*}{\rotatebox{90}{ETTm2}} 
&96 &\textbf{0.063} & \textbf{0.183} & \textbf{0.063} & 0.185 & 0.064 & 0.185 & 0.065 & 0.187 \\
&192 &\textbf{0.090} & \textbf{0.225} & 0.091 &0.227 & 0.093 & 0.228 & 0.093 & 0.230 \\
&336 &\textbf{0.118} & \textbf{0.262} & \textbf{0.118} & 0.263  & 0.119 & 0.263 & 0.122 & 0.267 \\
&720 &\textbf{0.172} & \textbf{0.320} & \textbf{0.172} & 0.322 & 0.173 & 0.323 & 0.173 & 0.324 \\

\midrule
\multirow{4}{*}{\rotatebox{90}{Electricity}} 
&96 &\textbf{0.194} & \textbf{0.304} & 0.196 & 0.306 & 0.198 & 0.309 & 0.209 & 0.321 \\
&192 &\textbf{0.226} & \textbf{0.325} & 0.229 & 0.330 & 0.231 & 0.330 & 0.246 & 0.344 \\
&336 &\textbf{0.263} & \textbf{0.358} & 0.269 & 0.363 & 0.271 & 0.364 & 0.280 & 0.370 \\
&720 &\textbf{0.298} & \textbf{0.404} & 0.317 & 0.413 & 0.322 & 0.417 & 0.332 & 0.424 \\

\midrule
\multirow{4}{*}{\rotatebox{90}{Weather}}
&96 &\textbf{0.00091} & \textbf{0.0210} & 0.00113 & 0.0252 & 0.00111 & 0.0248 & 0.00132 & 0.0265 \\
&192 &\textbf{0.00113} & \textbf{0.0239} & 0.00128 & 0.0262 & 0.00132 & 0.0275 & 0.00144 & 0.0281 \\
&336 &\textbf{0.00130} & \textbf{0.0258} & 0.00148 & 0.0279 & 0.00150 & 0.0294 & 0.00152 & 0.0289 \\
&720 &\textbf{0.00178} & \textbf{0.0305} & 0.00202 & 0.0328 & 0.00204 & 0.0331 & 0.00209 & 0.0341 \\

\midrule
\multirow{4}{*}{\rotatebox{90}{Traffic}}
&96 &\textbf{0.115} & \textbf{0.183} & 0.118 & 0.194 & 0.120 & 0.199 & 0.134 & 0.223 \\
&192 &\textbf{0.112} & \textbf{0.182} & 0.113 & 0.188 & 0.117 & 0.196 & 0.130 & 0.221 \\
&336 &\textbf{0.113} & \textbf{0.190} & 0.116 & 0.195 & 0.117 & 0.197 & 0.133 & 0.227 \\
&720 &\textbf{0.126} & \textbf{0.204} & 0.132 & 0.214 & 0.133 & 0.217 & 0.146 & 0.241 \\

\midrule
\bottomrule
\end{tabular}
\label{tab:appendix_modules}
}

\end{sc}
% \end{small}
\end{center}
\vskip 0.2in
% \end{adjustwidth}
%\end{footnotesize}
\end{table*}
%}

%% file: tables_vq/ablation_vq_boosting.tex
%\resizebox{!}{\.5\paperheight}{

\begin{table*}[h]
%\vskip -0.15in
\centering
%\begin{footnotesize}
% \begin{adjustwidth}{-1.5in}{-1in}

\caption{Results for boosting effect of sparse-VQ. We use FEDformer and Autoformer as backbones and leverage them with the Sparse-VQ. A lower MSE indicates better performance. All experiments are repeated 3 times.}%\vspace{-1mm}
\begin{center}
\begin{small}
\begin{sc}

\scalebox{1.0}{
\begin{tabular}{c|c|ccccccccccccccc}
\toprule
\multicolumn{2}{c|}{Methods}&\multicolumn{2}{c|}{FEDformer}&\multicolumn{2}{c|}{FEDformer+SVQ}&\multicolumn{2}{c|}{Autoformer}&\multicolumn{2}{c}{Autoformer+SVQ}\\
\midrule
\multicolumn{2}{c|}{Metric} & MSE  & MAE & MSE & MAE& MSE  & MAE& MSE & MAE\\
\midrule
\multirow{4}{*}{\rotatebox{90}{ETTm2}} 
&96 &0.203 & 0.287 & \textbf{0.192} & \textbf{0.286} & 0.255 & 0.339 & \textbf{0.214} & \textbf{0.295} \\
&192 &0.269 & \textbf{0.328} & \textbf{0.263} & \textbf{0.328} & 0.281 & 0.340 & \textbf{0.271} & \textbf{0.328} \\
&336 &\textbf{0.325} & 0.366 & \textbf{0.325} & \textbf{0.365} & 0.339 & 0.372 & \textbf{0.326} & \textbf{0.365} \\
&720 &\textbf{0.421} & \textbf{0.415} &0.432 & 0.424 & 0.422 & 0.419 & \textbf{0.412} & \textbf{0.410} \\

\midrule
\multirow{4}{*}{\rotatebox{90}{Electricity}} 
&96  &0.193 & 0.308 & \textbf{0.186} & \textbf{0.301} & 0.201 & 0.317 & \textbf{0.199} & \textbf{0.310} \\
&192  &0.201 & 0.315 & \textbf{0.197} & \textbf{0.311} & \textbf{0.222} & \textbf{0.334} & 0.235 & 0.341 \\
&336  &\textbf{0.214} & \textbf{0.329} & 0.218 & 0.333 & 0.231 & 0.338 & \textbf{0.226} & \textbf{0.337} \\
&720  &0.246 & 0.355 & \textbf{0.234} & \textbf{0.346} & \textbf{0.254} & 0.361 & 0.267 & \textbf{0.342} \\

\midrule
\multirow{4}{*}{\rotatebox{90}{Traffic}} 
&96  &0.587 & 0.366 & \textbf{0.569} & \textbf{0.354} & 0.613 & 0.388 & \textbf{0.595} & \textbf{0.366} \\
&192  &\textbf{0.604} & \textbf{0.373} & 0.611 & 0.378 & 0.616 & 0.382 & \textbf{0.592} & \textbf{0.365} \\
&336  &0.621 & 0.383 & \textbf{0.615} & \textbf{0.377} & 0.622 & 0.337 & \textbf{0.611} & \textbf{0.336} \\
&720  &\textbf{0.626} & \textbf{0.382} & 0.630 & 0.383 & \textbf{0.660} & \textbf{0.408} & 0.703 & 0.450 \\

\midrule
\multirow{4}{*}{\rotatebox{90}{Weather}}
&96 &0.217 & \textbf{0.296} & \textbf{0.209} & \textbf{0.296} & 0.266 & 0.336 & \textbf{0.224} & \textbf{0.299} \\
&192  &0.276 & 0.336 & \textbf{0.270} & \textbf{0.332} & 0.307 & 0.367 & \textbf{0.280} & \textbf{0.354} \\
&336  &0.339 & 0.38  & \textbf{0.319} & \textbf{0.375} & 0.359 & 0.395 & \textbf{0.330} & \textbf{0.372} \\
&720  &0.403 & 0.428 & \textbf{0.385} & \textbf{0.390} & 0.419 & 0.428 & \textbf{0.397} & \textbf{0.399} \\
\midrule
\bottomrule
\end{tabular}
\label{tab:ablation_vq_boosting}
}
\end{sc}
\end{small}
\end{center}
\vskip -0.1in
% \end{adjustwidth}
%\end{footnotesize}
\end{table*}
%}

%% file: tables_vq/different_vq_structure.tex
%\resizebox{!}{\.5\paperheight}{
% \vskip -0.2in
\begin{table*}[t]
\centering

%\begin{footnotesize}
% \begin{adjustwidth}{-1.5in}{-1in}
\caption{Univariate long-term series forecasting results of different sturcture of VQ. The best results are in bold. A lower MSE indicates better performance. All experiments are repeated 3 times.}\vspace{-1mm}

\begin{center}
% \begin{small}
\begin{sc}
\scalebox{0.80}{
\begin{tabular}{c|c|cccccccccccccccc}
\toprule
% \multicolumn{2}{c|}{Methods}&\multicolumn{2}{c|}{PatchTST(96)}&\multicolumn{2}{c|}{Attention}&\multicolumn{2}{c|}{Series-LG/GL}&\multicolumn{2}{c|}{Series-GL}&\multicolumn{2}{c}{Concatenate}\\
\multicolumn{2}{c|}{Methods}&\multicolumn{2}{c|}{$Sparse-VQ$}&\multicolumn{2}{c|}{$VQ$}&\multicolumn{2}{c|}{$VQ_{cosine}$}&\multicolumn{2}{c|}{$VQ_{kmeans}$}&\multicolumn{2}{c|}{$VQ_{recursive}$}&\multicolumn{2}{c}{$VQ_{AdaptiveCodebook}$}\\
\midrule
\multicolumn{2}{c|}{Metric} & MSE  & MAE & MSE & MAE& MSE  & MAE& MSE  & MAE& MSE  & MAE & MSE  & MAE\\
\midrule
\multirow{5}{*}{\rotatebox{90}{ECL}} 

 & 96  & 0.194          & 0.304          & 0.199  & 0.308   & 0.203  & 0.314  & 0.198           & 0.307     & 0.201   & 0.310    & 0.202          & 0.309       \\
 & 192 & 0.226          & 0.325          & 0.232  & 0.330    & 0.242  & 0.342  & 0.232           & 0.331     & 0.233   & 0.334     & 0.233          & 0.333    \\
 & 336 & 0.263          & 0.358          & 0.273  & 0.365    & 0.280  & 0.377  & 0.302           & 0.376       & 0.274   & 0.365  & 0.275          & 0.365     \\
 & 720 & 0.298          & 0.404          & 0.322  & 0.417    & 0.335  & 0.426  & 0.321           & 0.417     & 0.323   & 0.419     & 0.328          & 0.424      \\
 & Avg & \textbf{0.245} & \textbf{0.348} & 0.257  & 0.355   & 0.265  & 0.365  & 0.263           & 0.358      & 0.258   & 0.357    & 0.260          & 0.358      \\

\midrule
\multirow{5}{*}{\rotatebox{90}{Traffic}}
 & 96  & 0.115          & 0.183          & 0.116  & 0.186    & 0.119  & 0.193  & 0.117           & 0.186     & 0.118   & 0.190   & 0.115          & 0.182       \\
 & 192 & 0.112          & 0.182          & 0.113  & 0.185    & 0.118  & 0.194  & 0.113           & 0.185     & 0.116   & 0.191      & 0.111          & 0.184      \\
 & 336 & 0.113          & 0.190          & 0.113  & 0.194  & 0.118  & 0.201  & 0.115           & 0.190      & 0.116   & 0.196     & 0.114          & 0.192        \\
 & 720 & 0.126          & 0.204          & 0.129  & 0.208    & 0.136  & 0.221  & 0.133           & 0.211     & 0.137   & 0.220     & 0.127          & 0.205      \\
 & Avg & \textbf{0.117} & \textbf{0.190} & 0.118  & 0.193   & 0.123  & 0.202  & 0.120           & 0.193      & 0.122   & 0.199     & \textbf{0.117} & 0.191      \\

\midrule
\multirow{5}{*}{\rotatebox{90}{Weather}}
 & 96  & 0.0009         & 0.0210         & 0.0009 & 0.0212  & 0.0009 & 0.0218 & 0.0009          & 0.0211     & 0.0009  & 0.0214   & 0.0009         & 0.0212     \\
 & 192 & 0.0011         & 0.0239         & 0.0011 & 0.0235  & 0.0011 & 0.0240 & 0.0010          & 0.0235     & 0.0011  & 0.0238    & 0.0011         & 0.0233     \\
 & 336 & 0.0013         & 0.0258         & 0.0012 & 0.0255  & 0.0013 & 0.0261 & 0.0012          & 0.0255     & 0.0012  & 0.0258   & 0.0013         & 0.0257     \\
 & 720 & 0.0018         & 0.0305         & 0.0018 & 0.0304  & 0.0019 & 0.0316 & 0.0017          & 0.0303    & 0.0017  & 0.0302    & 0.0018         & 0.0305     \\
 & Avg & 0.0013         & \textbf{0.025} & 0.0013 & 0.0252   & 0.0013 & 0.0259 & \textbf{0.0012} & \textbf{0.025} & 0.0012 & 0.025   & 0.0013    & \textbf{0.025}   \\

\midrule
\bottomrule
\end{tabular}
\label{tab:different_vq_structure}
}

\end{sc}
% \end{small}
\end{center}
\vskip 0.15in
% \end{adjustwidth}
%\end{footnotesize}
\end{table*}
%}

%% file: tables_vq/ablation_noise_VQ.tex
%\resizebox{!}{\.5\paperheight}{

\begin{table*}[h]
%\vskip -0.15in
\centering
%\begin{footnotesize}
% \begin{adjustwidth}{-1.5in}{-1in}

\caption{Robustness analysis of univariate results conducted on four typical datasets. The degree of noise injected into the time series data is determined by $\eta$. A lower MSE indicates better performance. All experiments are repeated 3 times.}\vspace{-1mm}
\begin{center}
\begin{small}
\begin{sc}

\scalebox{1.0}{
\begin{tabular}{c|c|cccccccccccccccccc}
\toprule
\multicolumn{2}{c|}{Sparse-VQ}&\multicolumn{2}{c|}{Original}&\multicolumn{2}{c|}{$\eta=1\%$}&\multicolumn{2}{c|}{$\eta=5\%$}&\multicolumn{2}{c}{$\eta=10\%$}\\
\midrule
\multicolumn{2}{c|}{Metric} & MSE  & MAE & MSE & MAE& MSE  & MAE& MSE & MAE\\
\midrule
\multirow{4}{*}{\rotatebox{90}{ETTm2}} 
&96 &0.063 & 0.183 & 0.065 & 0.188 & 0.067 & 0.193 & 0.069 & 0.197 \\
&192 &0.090 & 0.225 & 0.091 & 0.228 & 0.093 & 0.231 & 0.094 & 0.233 \\
&336 &0.118 & 0.262 & 0.118  & 0.262 & 0.120  & 0.265 & 0.121  & 0.267 \\
&720 &0.172 & 0.322 & 0.173  & 0.323 & 0.174  & 0.323 & 0.173  & 0.323 \\

\midrule
\multirow{4}{*}{\rotatebox{90}{Electricity}} 
&96  &0.196 & 0.306 & 0.198 & 0.309 & 0.202 & 0.303 & 0.209 & 0.322 \\
&192 &0.229 & 0.330 & 0.230 & 0.331 & 0.233 & 0.334 & 0.241 & 0.345 \\
&336 &0.269 & 0.363 & 0.271 & 0.365 & 0.274 & 0.367 & 0.283 & 0.379 \\
&720 &0.317 & 0.413 & 0.316 & 0.415 & 0.322 & 0.420 & 0.328 & 0.425 \\

\midrule
\multirow{4}{*}{\rotatebox{90}{Traffic}} 
&96  &0.118 & 0.194 & 0.118 & 0.195 & 0.124 & 0.209 & 0.127 & 0.213 \\
&192  &0.115 & 0.192 & 0.116 & 0.194 & 0.120 & 0.203 & 0.136 & 0.240 \\
&336  &0.116 & 0.195 & 0.116 & 0.197 & 0.121 & 0.211 & 0.137 & 0.243 \\
&720  &0.132 & 0.214 & 0.132 & 0.217 & 0.136 & 0.226 & 0.153 & 0.259 \\

\midrule
\multirow{4}{*}{\rotatebox{90}{Weather}}
&96  &0.00113 & 0.0252 & 0.00114 & 0.0253 & 0.00116 & 0.0260 & 0.00115 & 0.0256 \\
&192  &0.00128 & 0.0262 & 0.00128 & 0.0262 & 0.00128 & 0.0262 & 0.00128 & 0.0263 \\
&336  &0.00148 & 0.0279 & 0.00148 & 0.0280 & 0.00147 & 0.0280 & 0.00152 & 0.0298 \\
&720  &0.00202 & 0.0328 & 0.00203 & 0.0328 & 0.00202 & 0.0326 & 0.00200 & 0.0329 \\

\midrule
\bottomrule

\end{tabular}

}
\label{tab:ablation_noise_VQ}
\end{sc}
\end{small}
\end{center}
\vskip 0.2in
% \end{adjustwidth}
%\end{footnotesize}
\end{table*}
%}

%% file: appendix_vq/appendix_noise_PatchTST.tex
%\resizebox{!}{\.5\paperheight}{

\begin{table*}[h]
%\vskip -0.2in
\centering
%\begin{footnotesize}
% \begin{adjustwidth}{-1.5in}{-1in}

\caption{Robustness analysis of unitivariate results conducted on four typical datasets for PatchTST. The degree of noise injected into the time series data is determined by $\eta$. A lower MSE indicates better performance. All experiments are repeated 3 times.}\vspace{-1mm}
\begin{center}
\begin{small}
\begin{sc}

\scalebox{1.0}{
\begin{tabular}{c|c|cccccccccccccccccc}
\toprule
\multicolumn{2}{c|}{PatchTST}&\multicolumn{2}{c|}{Original}&\multicolumn{2}{c|}{$\eta=1\%$}&\multicolumn{2}{c|}{$\eta=5\%$}&\multicolumn{2}{c}{$\eta=10\%$}\\
\midrule
\multicolumn{2}{c|}{Metric} & MSE  & MAE & MSE & MAE& MSE  & MAE& MSE & MAE\\
\midrule
\multirow{4}{*}{\rotatebox{90}{ETTm2}} 
&96 &0.0648 & 0.1869 & 0.065  & 0.189 & 0.069  & 0.196 & 0.074  & 0.204 \\
&192 &0.0929 & 0.2304 & 0.0945 & 0.233 & 0.0973 & 0.237 & 0.0981 & 0.239 \\
&336 &0.1218 & 0.2672 & 0.122  & 0.269 & 0.124  & 0.269 & 0.126  & 0.274 \\
&720 &0.1733 & 0.3238 & 0.172  & 0.322 & 0.177  & 0.327 & 0.182  & 0.333 \\

\midrule
\multirow{4}{*}{\rotatebox{90}{Electricity}} 
&96  &0.209 & 0.321 & 0.206 & 0.315 & 0.211 & 0.325 & 0.224 & 0.340 \\
&192  &0.246 & 0.344 & 0.243 & 0.343 & 0.249 & 0.350 & 0.266 & 0.369 \\
&336  &0.280 & 0.370 & 0.282 & 0.373 & 0.287 & 0.381 & 0.302 & 0.395 \\
&720  &0.332 & 0.424 & 0.330 & 0.426 & 0.338 & 0.433 & 0.355 & 0.446 \\

\midrule
\multirow{4}{*}{\rotatebox{90}{Traffic}} 
&96  &0.134 & 0.223 & 0.134 & 0.226 & 0.157 & 0.253 & 0.161 & 0.256 \\
&192  &0.130 & 0.221 & 0.138 & 0.229 & 0.142 & 0.239 & 0.156 & 0.251 \\
&336  &0.133 & 0.227 & 0.136 & 0.232 & 0.150 & 0.253 & 0.163 & 0.263 \\
&720  &0.146 & 0.241 & 0.150 & 0.245 & 0.165 & 0.264 & 0.182 & 0.281 \\

\midrule
\multirow{4}{*}{\rotatebox{90}{Weather}}
&96  &0.00132 & 0.0265 & 0.00123 & 0.0256 & 0.00133 & 0.0267 & 0.00130 & 0.0264 \\
&192  &0.00144 & 0.0281 & 0.00137 & 0.0274 & 0.00140 & 0.0277 & 0.00143 & 0.0281 \\
&336  &0.00152 & 0.0289 & 0.00153 & 0.0291 & 0.00151 & 0.0287 & 0.00154 & 0.0291 \\
&720  &0.00209 & 0.0341 & 0.00206 & 0.0339 & 0.00201 & 0.0331 & 0.00204 & 0.0335 \\

\bottomrule
\end{tabular}
\label{tab:appendix_noise_PatchTST}
}
\end{sc}
\end{small}
\end{center}
\vskip 0.15in
% \end{adjustwidth}
%\end{footnotesize}
\end{table*}
%}

%% file: appendix_vq/appendix_noise_FEDformer.tex
%\resizebox{!}{\.5\paperheight}{

\begin{table*}[h]
%\vskip -0.2in
\centering
%\begin{footnotesize}
% \begin{adjustwidth}{-1.5in}{-1in}

\caption{Robustness analysis of unitivariate results conducted on four typical datasets for FEDformer. The degree of noise injected into the time series data is determined by $\eta$. A lower MSE indicates better performance. All experiments are repeated 3 times.}\vspace{-1mm}
\begin{center}
\begin{small}
\begin{sc}

\scalebox{1.0}{
\begin{tabular}{c|c|cccccccccccccccccc}
\toprule
\multicolumn{2}{c|}{FEDformer}&\multicolumn{2}{c|}{Original}&\multicolumn{2}{c|}{$\eta=1\%$}&\multicolumn{2}{c|}{$\eta=5\%$}&\multicolumn{2}{c}{$\eta=10\%$}\\
\midrule
\multicolumn{2}{c|}{Metric} & MSE  & MAE & MSE & MAE& MSE  & MAE& MSE & MAE\\
\midrule
\multirow{4}{*}{\rotatebox{90}{ETTm2}} 
&96 &0.0648 & 0.1869 & 0.065  & 0.189 & 0.069  & 0.196 & 0.074  & 0.204 \\
&192 &0.0929 & 0.2304 & 0.0945 & 0.233 & 0.0973 & 0.237 & 0.0981 & 0.239 \\
&336 &0.1218 & 0.2672 & 0.122  & 0.269 & 0.124  & 0.269 & 0.126  & 0.274 \\
&720 &0.1733 & 0.3238 & 0.172  & 0.322 & 0.177  & 0.327 & 0.182  & 0.333 \\

\midrule
\multirow{4}{*}{\rotatebox{90}{Electricity}} 
&96  &0.209 & 0.321 & 0.206 & 0.315 & 0.211 & 0.325 & 0.224 & 0.340 \\
&192  &0.246 & 0.344 & 0.243 & 0.343 & 0.249 & 0.350 & 0.266 & 0.369 \\
&336  &0.280 & 0.370 & 0.282 & 0.373 & 0.287 & 0.381 & 0.302 & 0.395 \\
&720  &0.332 & 0.424 & 0.330 & 0.426 & 0.338 & 0.433 & 0.355 & 0.446 \\

\midrule
\multirow{4}{*}{\rotatebox{90}{Traffic}} 
&96  &0.134 & 0.223 & 0.134 & 0.226 & 0.157 & 0.253 & 0.161 & 0.256 \\
&192  &0.130 & 0.221 & 0.138 & 0.229 & 0.142 & 0.239 & 0.156 & 0.251 \\
&336  &0.133 & 0.227 & 0.136 & 0.232 & 0.150 & 0.253 & 0.163 & 0.263 \\
&720  &0.146 & 0.241 & 0.150 & 0.245 & 0.165 & 0.264 & 0.182 & 0.281 \\

\midrule
\multirow{4}{*}{\rotatebox{90}{Weather}}
&96  &0.00132 & 0.0265 & 0.00123 & 0.0256 & 0.00133 & 0.0267 & 0.00130 & 0.0264 \\
&192  &0.00144 & 0.0281 & 0.00137 & 0.0274 & 0.00140 & 0.0277 & 0.00143 & 0.0281 \\
&336  &0.00152 & 0.0289 & 0.00153 & 0.0291 & 0.00151 & 0.0287 & 0.00154 & 0.0291 \\
&720  &0.00209 & 0.0341 & 0.00206 & 0.0339 & 0.00201 & 0.0331 & 0.00204 & 0.0335 \\

\bottomrule
\end{tabular}
\label{tab:appendix_noise_FEDformer}
}
\end{sc}
\end{small}
\end{center}
%\vskip -0.1in
% \end{adjustwidth}
%\end{footnotesize}
\end{table*}
%}

%% file: appendix_vq/appendix_proof.tex
\section{Detailed Proof}
\label{app:proof}
\begin{proof}
\renewcommand{\qedsymbol}{}
To establish an upper limit for $N(\mathcal{U}, \epsilon)$ using clustering, the covering number for a unit sphere $\mathcal{U}$ is considered, which necessitates at least $1/\epsilon^n$ codewords for approximate representation within an error of $\epsilon$. Let $Vf$, where $V = (v_1, \ldots, v_l)$, $v_j \sim \mathcal{N}(0, I_n/l)$, and $f \in \Omega_t$ is a $t$-sparse unit vector. We have:
\[
\Pr\left(\|VV^{\top} - I\|_2 \geq \lambda \right) \leq 2n\exp\left(-\frac{l\lambda^2}{3n}\right),
\]
which implies
\[
\|VV^{\top} - I\|_2 \leq \Gamma := \sqrt{\frac{n}{l}\log\frac{2n}{\eta}}
\]
with a probability of at least $1 - \eta$. Consequently, $\|f' - f\|_2 \geq (1+\Gamma)^{-1} \|Vf - Vf'\|_2$. Given the $t$-sparse unit vector covering number is capped by $(Dl/t\epsilon)^t$, we deduce:
\[
\left(\frac{Dl}{t\epsilon}\right)^t \geq \left(1 + \frac{2}{\epsilon}\right)^n(1+\Gamma)^n,
\]
Selecting $l = (4n/\epsilon)^q$ ensures:
\[
(1+\Gamma)^n \leq \exp\left(n\Gamma\right) \leq e,
\]
which results in $t\log(D'l/\epsilon) \geq 2n/\epsilon + t\log t$, with $D' = De$. 
With the assumption that $t \geq 4n/[\epsilon(\log K + q\log(4n) - (q+1)\log\epsilon)]$, it holds that $t\log t \leq 2n/\epsilon$, thereby concluding that $l \geq (4n/\epsilon)^q$.
\end{proof}